\documentclass[acmsmall,screen,nonacm]{acmart}

\AtBeginDocument{%
  }

\usepackage{tcolorbox}
\usepackage{algpseudocode}
\usepackage{multirow}
\usepackage{booktabs}
\usepackage{enumitem}
\usepackage{xcolor}
\usepackage{wrapfig}
\usepackage{subcaption}
\usepackage[normalem]{ulem}

\setcopyright{acmcopyright}
\copyrightyear{2024}
\acmYear{2024}
\acmDOI{XXXXXXX.XXXXXXX}
\acmConference[POPL]{2024}{London, England}

\acmBooktitle{Woodstock '18: ACM Symposium on Neural Gaze Detection, June 03--05, 2024, Woodstock, NY}

\newcommand{\tool}{\texttt{NeuralSAT}}
\newcommand{\crown}{\texttt{$\alpha$-$\beta$-CROWN}}
\newcommand{\nnenum}{\texttt{nnenum}}
\newcommand{\marabou}{\texttt{Marabou}}
\newcommand{\marabouold}{\texttt{Marabou'21}}
\newcommand{\marabounew}{\texttt{Marabou'22}}
\newcommand{\eran}{\texttt{ERAN}}
\newcommand{\reluplex}{\texttt{Reluplex}}
\newcommand{\reluval}{\texttt{Reluval}}
\newcommand{\neurify}{\texttt{Neurify}}
\newcommand{\nnv}{\texttt{NNV}}
\newcommand{\dnnv}{\texttt{DNNV}}
\newcommand{\verinet}{\texttt{VeriNet}}
\newcommand{\mnbab}{\texttt{MN-BaB}}

\newcommand{\vnncomp}{VNN-COMP'22}
\newcommand{\planet}{\texttt{Planet}}

\newtheorem{theorem}{Theorem}[section]

\newtheorem{lemma}[theorem]{Lemma}

\usepackage[vlined,figure,linesnumbered]{algorithm2e}

\SetCommentSty{haicommentfont}

\newtoggle{usecomment}
\settoggle{usecomment}{true}

\newcommand{\tvn}[1]{\iftoggle{usecomment}{{\color{red}{[TVN]: #1}}}{}}
\newcommand{\hd}[1]{\iftoggle{usecomment}{{\color{blue}{[HD]: #1}}}{}}

\newcommand{\ignore}[1]{}

\begin{document}

\title{A DPLL(T) Framework for Verifying Deep Neural Networks}

\author{Hai Duong}
\email{hduong22@gmu.edu}
\affiliation{%
  \institution{George Mason University}
  \country{USA}
}

\author{ThanhVu Nguyen}
\email{tvn@gmu.edu}
\affiliation{%
  \institution{George Mason University}
  \country{USA}
}
\author{Matthew B. Dwyer}
\email{matthewbdwyer@virginia.edu}
\affiliation{
\institution{University of Virginia}
\country{USA}
}


\begin{abstract}
  Deep Neural Networks (DNNs) have emerged as an effective approach to tackling real-world problems.
  However, like human-written software, DNNs can have bugs and can be attacked. To address this, research has explored a wide-range of algorithmic approaches to verify DNN behavior.
  In this work, we introduce \tool{}, a new verification approach that adapts the widely-used DPLL(T) algorithm used in modern SMT solvers.  A key feature of SMT solvers is the use of conflict clause learning and search restart to scale verification.  Unlike prior DNN verification approaches, \tool{} combines an abstraction-based deductive theory solver with clause learning and an evaluation clearly demonstrates the benefits of the approach on a set of challenging verification benchmarks.
\end{abstract}


\maketitle

    
    
    

\section{Introduction}

Deep Neural Networks (DNNs) have emerged as an effective approach for solving challenging real-world problems. However, just like traditional software, DNNs can have ``bugs'', e.g., producing unexpected results on inputs that are different from those in training data, and be attacked, e.g., small perturbations to the inputs by a malicious adversary or even sensor imperfections can result in misclassification~\cite{Ren2020, Zugner2019, yang2022natural,zhang2019empirical,isac2022neural}. 
These issues, which have been observed in many DNNs~\cite{goodfellow2014explaining,szegedy2014intriguing} and demonstrated in the real world~\cite{eykholt2018robust},  naturally raise the question of how DNNs should be tested, validated, and ultimately \emph{verified}
to meet the requirements of relevant robustness or safety standards~\cite{huang2020survey,katz2017towards}.

To address this question, researchers have developed a wide-variety of algorithmic techniques and supporting tools to verify properties of DNNs (e.g.,~\cite{katz2017reluplex,ehlers2017formal,huang2017safety,katz2019marabou,wang2018formal,singh2018fast,singh2019abstract,katz2022reluplex,urban2021review,liu2021algorithms,muller2021scaling,wang2021beta}).  
Recent instances of the DNN verification tool competition (VNN-COMP) indicate that three key elements
 are common to the best performing approaches:
(1) the use of abstraction to reason symbolically about sets of neuron output values;
(2) the use of neuron splitting to specialize the analysis of subproblems in a form of branch-and-bound (BaB) reasoning; and
(3) the use of fast-path optimizations that can discharge easy verification problems quickly~\cite{bak2021second,muller2022third}.
For example, the top four performers in VNN-COMP 2022: \crown{}~\cite{wang2021beta,zhang2022general}, \mnbab{}~\cite{ferrari2022complete}, \verinet{}~\cite{henriksen2020efficient}, and \nnenum{}~\cite{bak2021nnenum}, all include these features.

The problem of verifying non-trivial properties of DNNs with piecewise linear activation functions, such as ``ReLU'', has been shown to be reducible~\cite{katz2017reluplex} to the Boolean satisfiability (SAT) problem~\cite{cook1971complexity}.
Thus, at its core, any DNN verification algorithm must contend with worst-case exponential complexity.
As the fields of SAT and satisfiability modulo theory (SMT) solving have demonstrated, despite this
complexity well-chosen combinations
of algorithmic techniques can solve a wide-range of large real-world problems~\cite{kroening2016decision}.
In this paper, we explore the design of an SMT-inspired DPLL(T) solver customized for DNN verification that is
competitive with the state-of-the-art and that establishes a foundation for incorporating additional algorithmic
techniques from the broader SMT literature.

We are not the first to explore SMT solving for DNN verification.
The earliest techniques in the field, \planet{}~\cite{ehlers2017formal} and \reluplex{}~\cite{katz2017reluplex}, demonstrated how the semantics of a trained DNN could be encoded as a constraint in Linear Real Arithmetic (LRA).
In principle, such constraints can be solved by any SMT solver equipped with an LRA
\textit{theory solver} (T-solver)~\cite{kroening2016decision}.
The DPLL(T) algorithm implemented by modern SMT solvers works by moving back and forth between solving an abstract propositional encoding of the constraint and solving a theory-specific encoding of a constraint fragment corresponding
to a partial assignment of propositional literals.  
The challenge in solving DNN verification constraints
lies in the fact that each neuron gives rise to a disjunctive constraint to encode its non-linear behavior.   
In practice, this leads to a combinatorial blowup in the space of assignments the SMT solver must consider at the abstract propositional level.
To resolve the exponential complexity inherent in such constraints, both \planet{} and \reluplex{} chose to \textit{push} the disjunctive constraints from the propositional encoding down into the theory-specific encoding of the problem, leveraging a technique referred to as splitting-on-demand~\cite{barrett2006splitting}.
This works to an extent, but it does not scale well to large DNNs~\cite{bak2021second,muller2022third}.
We observe that the choice to pursue an aggressive splitting-on-demand strategy
sacrifices the benefit of several of the key algorithmic techniques that make SMT solvers scale -- specifically conflict-driven clause learning (CDCL)~\cite{bayardo1997using,marques1999grasp,569607}, theory propagation~\cite{kroening2016decision},
and search restart~\cite{pipatsrisawat2009power}.  

We present the \textbf{\tool{}} framework, which consists of a lazy, incremental LRA-solver that is parameterized by state-of-the-art abstractions, such as LiRPA~\cite{xu2020fast,wang2021beta}, to efficiently perform deductive reasoning and exhaustive theory propagation~\cite{nieuwenhuis2006solving}, and to support restarts.
As prior research has demonstrated~\cite{nieuwenhuis2006solving}, the interplay
between CDCL and restart is essential to scaling and we find that it permits \tool{} to increase the number of problems verified by 53\% on a challenging benchmark (\S\ref{sec:ablation}).
Moreover, \tool{} significantly outperforms the best existing DPLL-based DNN verification approach -- \marabou{}~\cite{katz2019marabou,katz2022reluplex} -- which also employs abstraction and deduction, but does not exploit clause learning (\S\ref{sec:comparison-marabou}).  Moreover, despite the fact that \tool{} is an early stage
prototype, that does not incorporate the fast-path optimizations of other tools, it ranks second to \crown{} in solving benchmarks from the VNN-COMP competition (\S\ref{sec:vnncomp}).

The contributions of this work lie in:
(i) developing a domain-specific LRA-solver that allows for the benefits of clause learning to accelerate SMT-based DNN verification;
(ii) developing a prototype \tool{} implementation which we release as open source; and
(iii) empirically demonstrating that the approach compares favorably with the state-of-the-art in terms of scalability, performance, and ability to solve challenging DNN verification problems.
These findings collectively suggest that techniques like CDCL are advantageous, in combination with other optimizations, in scaling DNN verification to larger networks.

\section{Background}\label{sec:background}

\subsection{Satisfiability and DPLL(T)}\label{sec:constraintsolving}


The classical satisfiability (SAT) problem asks if a given propositional formula over Boolean variables can be satisfied~\cite{biere2009handbook}.  Given a formula $f$, a SAT solver returns \texttt{sat} if it can find a satisfying assignment that maps truth values to variables of $f$ that makes $f$ evaluate to true, and \texttt{unsat} if it cannot find any satisfying assignments. The problem is NP-Complete and research into methods for efficiently solving  problem instances has been ongoing for multiple decades.

\begin{wrapfigure}{r}{0.40\linewidth}
\vspace{-0.1in}
  \includegraphics[width=1\linewidth]{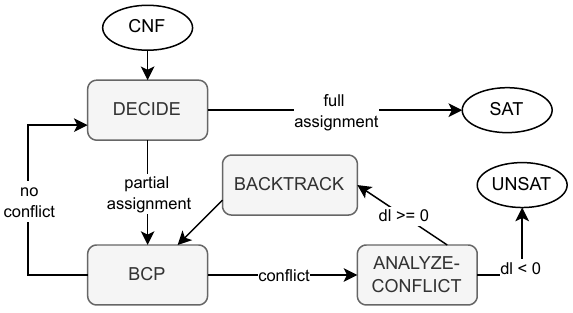}
  \caption{\label{fig:dpll} Original DPLL Algorithm.}
\vspace{-0.1in}  
\end{wrapfigure}

\paragraph{DPLL} Fig.~\ref{fig:dpll} gives an overview of \textbf{DPLL}, a SAT solving technique introduced in 1961 by Davis, Putnam, Logemann, and Loveland~\cite{davis1962machine}. DPLL is an iterative algorithm that takes as input a propositional formula and (i) decides an unassigned variable and assigns it a truth value, (ii) performs Boolean constraint propagation (BCP or also called Unit Propagation),  which detects single literal clauses that either force a literal to be true in a satisfying assignment or give rise to a conflict; (iii) analyzes the conflict to backtrack to a previous decision level \texttt{dl}; and (iv) erases assignments at levels greater than \texttt{dl} to try new  assignments. These steps repeat until DPLL finds  a satisfying assignment and returns \texttt{sat}, or decides that it cannot backtrack (\texttt{dl}=-1) and returns \texttt{unsat}.


Modern DPLL solving improves the original version with Conflict-Driven Clause Learning  (\textbf{CDCL})~\cite{bayardo1997using,marques1999grasp,569607}.
DPLL with CDCL can \emph{learn new clauses} to avoid past conflicts and backtrack more intelligently (e.g., using non-chronologically backjumping).
Due to its ability to learn new clauses, CDCL can significantly  reduce the search space and allow SAT solvers to scale to large problems.
In the following, whenever we refer to DPLL, we mean DPLL with CDCL.

\paragraph{DPLL(T)} DPLL(T)~\cite{nieuwenhuis2006solving} extends DPLL for propositional formulae to check SMT formulae involving non-Boolean variables, e.g., real numbers and data structures such as strings, arrays, lists.
DPLL(T) combines DPLL with dedicated \emph{theory solvers} to analyze formulae in those theories\footnote{SMT is Satisfiability Modulo Theories and the T in DPLL(T) stands for Theories.}.  
For example, to check a formula involving linear arithmetic over the reals (LRA), DPLL(T) may use a theory solver that uses linear programming to check the  constraints in the formula.
Modern DPLL(T)-based SMT solvers such as Z3~\cite{moura2008z3} and CVC4~\cite{barrett2011cvc4} 
include solvers supporting a wide range of theories including linear arithmetic, nonlinear arithmetic, string, and arrays~\cite{kroening2016decision}.



\subsection{The DNN verification problem}\label{sec:nnverif}

A \emph{neural network} (\textbf{NN})~\cite{Goodfellow-et-al-2016} consists of an input layer, multiple hidden layers, and an output layer. Each layer has a number of neurons, each connected to neurons from previous layers through a predefined set of weights (derived by training the network with data). A \textbf{DNN} is an NN with at least two hidden layers. 


The output of a DNN is obtained by iteratively computing  the  values  of  neurons  in  each  layer.
The value of a neuron in the input layer is the input data. The value of a neuron in the hidden layers is computed by applying an \emph{affine transformation} to values of neurons in the previous layers, then followed by an \emph{activation function} such as the popular Rectified Linear Unit (ReLU) activation.

For this activation, the value of a hidden neuron \(y\) is 
$ReLU(w_1v_1 + \dots{} + w_nv_n + b)$, where \(b\) is the bias parameter of \(y\), \(w_i, \dots, w_n\) are the weights of \(y\), \(v_1,\dots,v_n\) are the neuron values of preceding layer, \(w_1v_1 + \dots + w_nv_n+b\) is the affine transformation, and \(ReLU(x) = \max(x,0)\) is the ReLU activation. The values of a neuron in the output layer is evaluated similarly but it may skip the activation function.
A ReLU activated neuron is said to be \emph{active} if its input value is greater than
zero and \emph{inactive} otherwise.

\paragraph{DNN Verification} Given a DNN \(N\) and a property $\phi$, the \emph{DNN verification problem} asks if $\phi$ is a valid property of $N$.
Typically, $\phi$ is a formula of the form $\phi_{in} \Rightarrow \phi_{out}$, where $\phi_{in}$ is a property over the inputs of $N$ and $\phi_{out}$ is a property over the outputs of $N$.
A DNN verifier attempts to find a \emph{counterexample} input to $N$ that satisfies $\phi_{in}$ but violates $\phi_{out}$.  If no such counterexample exists, $\phi$ is a valid property of $N$. Otherwise, $\phi$ is not valid and the counterexample can be used to retrain or debug the DNN~\cite{huang2017safety}.



\begin{wrapfigure}{r}{0.3\textwidth}
    \centering
    \includegraphics[width=1\linewidth]{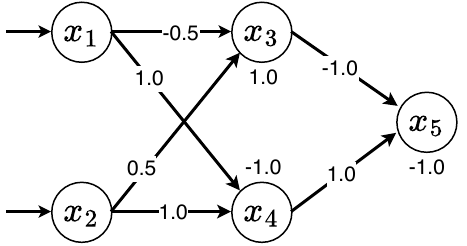}
    \caption{\label{fig:dnn} An FNN with ReLU.}
  \end{wrapfigure}
\paragraph{Example} Fig.~\ref{fig:dnn} shows a simple DNN with two inputs $x_1,x_2$, two hidden neurons $x_3,x_4$, and one output $x_5$. The weights of a neuron are shown on its incoming edges , and the bias is shown above or below each neuron. The outputs of the hidden neurons  are computed the affine transformation and ReLU, e.g., $x_3 = ReLU(-0.5x_1+0.5x_2+1.0)$. The output neuron is computed with just the affine transformation, i.e., $x_5=-x_3+x_4-1$.

A valid property for this DNN is that the output is $x_5 \le 0$ for any inputs $x_1 \in [-1,1], x_2\in[-2,2]$. An invalid property for this network is that $x_5 > 0$ for those similar inputs.
A counterexample showing this property violation is $\{x_1=-1, x_2=2\}$, from which the network evaluates to $x_5=-3.5$. Such properties can capture \emph{safety requirements} (e.g., a rule in an  collision avoidance system in~\cite{kochenderfer2012next,katz2017reluplex} is ``if the intruder is distant and significantly slower than us, then we stay below a certain threshold'') or \emph{local robustness}~\cite{katz2017towards} conditions (a form of adversarial robustness stating that small perturbations of a given input all yield the same output).


\paragraph{Abstraction}
ReLU-based DNN verification is NP-Complete~\cite{katz2017reluplex} and thus can be formulated as a SAT or SMT checking problem.
Direct application of SMT solvers does not scale to the large and complex formulae encoding real-world, complex DNNs.
While custom theory solvers, like \planet{} and \reluplex{}, retain the soundness, completeness,
and termination of SMT and improve on the performance of a direct SMT encoding, they too do not scale sufficiently to handle realistic DNNs~\cite{bak2021second,brix2023first}. 

Applying techniques from abstract interpretation~\cite{cousot1977abstract},
abstraction-based DNN verifiers overapproximate nonlinear computations (e.g., ReLU) of the network using linear abstract domains such as interval~\cite{wang2018formal}, zonotope~\cite{singh2018fast}, polytope~\cite{singh2019abstract,xu2020fast}. 
As illustrated in Fig.~\ref{fig:abs} abstract domains can model nonlinearity with
varying degrees of precision using polyhedra that are efficient to compute with.
This allows abstraction-based DNN verifiers to side-step the disjunctive splitting that is the performance bottleneck
of constraint-based DNN verifiers.

A DNN verification technique using an approximation, e.g., the polytope abstract domain, 
works by (i) representing the input ranges of the DNN as polytopes, (ii) applying transformation rules to the affine and ReLU computations of the network to compute polytope regions representing values of neurons, and (iii) finally, converting the polytope results into output bounds.
The resulting outputs are an overapproximation of the actual outputs.

\section{Overview of \tool{}}\label{sec:overview}

\begin{wrapfigure}{r}{0.3\textwidth}
\vspace{-0.15in}
    \centering
    \includegraphics[width=1\linewidth]{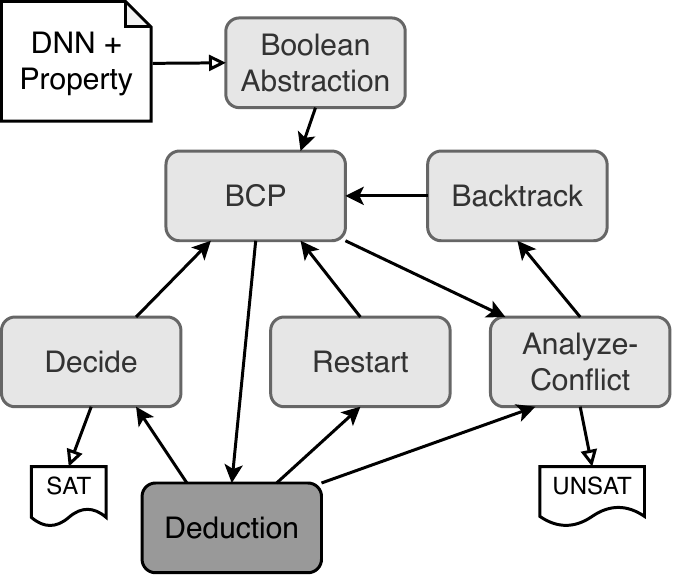}
    \caption{\label{fig:overview} \tool{}.} 
\end{wrapfigure}
\tool{} is a SMT-based DNN verifier that uses abstraction in its theory solver to  accelerate  unsatisfiability checking and the exploration of the space of variable assignments. 
Fig.~\ref{fig:overview} gives an overview of \tool{}, which follows the DPLL(T) framework (\S\ref{sec:background}) with some modification  and consists of standard DPLL components (light shades) and the theory solver (dark shade).

\tool{} constructs a propositional formula over Boolean variables that represent the activation status of neurons (\emph{Boolean Abstraction}).   Clauses in the formula assert that each neuron, e.g., neuron $i$, is active or inactive, e.g., $v_i \vee \overline{v_i}$.
This abstraction allows us to  use standard DPLL components to search for truth values satisfying these clauses and a DNN-specific theory solver to check the feasibility of truth assignments with respect to the constraints encoding the DNN and the property of interest.

\tool{} now enters an iterative process to find assignments satisfying the activation clauses.
First, \tool{} assigns a truth value to an unassigned variable (\emph{Decide}), detects unit clauses caused by this assignment, and infers additional assignments (\emph{Boolean Constraint Propagation}).
Next, \tool{} invokes the theory solver or T-solver (\emph{Deduction}), which uses LP solving and abstraction to check the satisfiability of the constraints of the current assignment with the property of interest. The T-solver can also infer additional truth assignments.

If the T-solver confirms satisfiability, \tool{} continues with new assignments (\emph{Decide}). Otherwise, \tool{} detects a conflict   (\emph{Analyze Conflict}) and learns clauses to remember it and backtrack to a previous decision  (\emph{Backtrack}).
If \tool{} detects local optima, it would restart (\emph{Restart}) the search by clearing all decisions that have been made to escape and the conflict clauses learned so far would be also recorded to avoid reaching the same state in the next runs.
As we discuss later in \S\ref{sec:eval}, restarting especially benefits challenging DNN problems by enabling better clause learning and exploring different decision orderings.
This process repeats until \tool{} can no longer backtrack, and return \texttt{unsat}, indicating the DNN has the property, or it finds a total assignment for all boolean variables, and returns \texttt{sat} (and the user can query \tool{} for a counterexample).

\subsection{Illustration}\label{sec:unsat}

We use \tool{} to prove that for inputs $x_1 \in [-1, 1], x_2 \in [-2,2]$ the DNN in Fig.~\ref{fig:dnn} produces the output $x_5 \le 0$.
\tool{} takes as input the formula $\alpha$ representing the DNN:
\begin{equation}\label{eq:ex}
\begin{aligned}
  x_3 = ReLU(-0.5x_1 + 0.5x_2 + 1)\;\;\;\land\;\;\;
  x_4 = ReLU(x_1 + x_2 - 1) \;\;\;\land\; \;\;
  x_5 = -x_3 + x_4 -1
\end{aligned}
\end{equation}
and the formula $\phi$ representing the property:
\begin{equation}\label{eq:valid_prop}
    \phi : -1\le x_1 \le 1 \land -2 \le x_2 \le 2 \quad\Rightarrow\quad x_5 \le 0.
  \end{equation}
To prove $\alpha \Rightarrow \phi$, \tool{} shows  that \emph{no} values of $x_1,x_2$ satisfying the input properties would result in $x_5 > 0$. Thus, we want \tool{} to return \texttt{unsat} for  $\overline{\alpha \Rightarrow \phi}$: 
\begin{equation}\label{eq:negprop}
  \alpha\; \land\; -1 \le x_1 \le 1     \;\land\; -2 \le x_2 \le 2   \;\land\; x_5 > 0.
\end{equation} 

In the following, we write $x \mapsto v$ to denote that the variable $x$ is assigned with a truth value $v \in \{T,F\}$. This assignment can be either decided by \texttt{Decide} or inferred by \texttt{BCP}. We also write $x@dl$ and  $\overline{x}@dl$ to indicate the respective assignments $x \mapsto T$ and $x \mapsto F$  at decision level $dl$.

\paragraph{Boolean Abstraction} First, \tool{} creates two Boolean variables $v_3$ and $v_4$ to represent the
activation status of the hidden neurons $x_3$ and $x_4$, respectively. For example, $v_3=T$ means $x_3$ is \texttt{active} and thus is the constraint $-0.5x_1 + 0.5x_2 + 1 > 0$. Similarly, $v_3=F$ means $x_3$ is \texttt{inactive} and therefore is $-0.5x_1 + 0.5x_2 + 1\le 0$. Next, \tool{} forms two clauses  $\{v_3 \lor \overline{v_3} \;;\; v_4 \lor \overline{v_4}\}$ indicating these variables are either \texttt{active} or \texttt{inactive}.

\paragraph{DPLL(T) Iterations} \tool{} searches for an assignment to satisfy the clauses and the constraints they represent.
For this example, \tool{} uses four iterations, summarized in Tab.~\ref{tab:valid}, to determine that no such assignment exists and the problem is thus \texttt{unsat}.

\begin{table*}
    \centering
    \caption{\tool{}'s run producing \texttt{unsat}.}\label{tab:valid}    
    \footnotesize
      \begin{tabular}{ccccccc}
      \toprule
      Iter & \textbf{BCP} & \multicolumn{2}{c}{\textbf{DEDUCTION}}& \textbf{DECIDE} & \multicolumn{2}{c}{\textbf{ANALYZE-CONFLICT}} \\
        &&Constraints&Bounds&&Bt&Learned Clauses\\
        \midrule
        Init &-& $I = -1 \le x_1 \le 1; -2 \le x_2 \le 2$ & $-1 \le x_1 \le 1; -2 \le x_2\le 2$ & - &-&$C = \{v_3 \lor \overline{v_3};\; v_4 \lor \overline{v_4}\}$\\
        
        1 &-&$I$ & $ x_5 \le 1 $& $\overline{v_4}@1$&-&-\\
        
        2 &-&$I; x_4=\texttt{off}$&$ x_5 \le -1$& - & 0 &  $C = C \cup \{v_4\}$\\
        
        3 &$v_4@0$&$I; x_4=\texttt{on} $&$ x_3 \ge 0.5; x_5 \le 0.5$ & $v_3@0$&-&-\\
        
        4 &-&$I; x_3=\texttt{on}; x_4=\texttt{on}$&-&- & \bf{-1} & $C = C\cup \{\overline{v_4}\}$\\
        
        
         \bottomrule
      \end{tabular}
  \end{table*}


In \emph{iteration 1}, as shown in Fig.~\ref{fig:overview}, \tool{} starts with \texttt{BCP}, which has no effects because the current clauses and (empty) assignment produce no unit clauses.
In \texttt{Deduction}, \tool{} uses an LP solver to determine that the current set of constraints, which contains just the initial input bounds, is feasible\footnote{We use the terms feasible, from the LP community, and satisfiable, from the SAT community, interchangeably.}. \tool{} then uses abstraction to approximate an output upper bound $x_5 \le 1$ and thus deduces that satisfying the output $x_5 >0$ might be feasible. \tool{} continues with \texttt{Decide}, which uses a heuristic to select the unassigned variable $v_4$ and sets $v_4=F$.  \tool{} also increments the decision level ($dl$) to 1 and associates $dl=1$ to the assignment, i.e., $\overline{v_4}@1$. Note that this process of selecting and assigning (random) values to variables representing neurons is commonly called \emph{neuron splitting} because it splits the search tree into subtrees corresponding into the assigned values (e.g., see \S\ref{sec:restart-tree}).

In \emph{iteration 2}, \texttt{BCP} again has no effect because it does not detect any unit clauses. In \texttt{Deduction}, \tool{} determines that current set of constraints, which contains $x_1 + x_2 - 1 \le 0$ due to the assignment $v_4\mapsto F$ (i.e., $x_4=\texttt{off}$), is feasible. \tool{} then approximates a new output upper bound $x_5\le -1$, which means satisfying the output $x_5 > 0$ constraint is \emph{infeasible}.

\tool{} now enters \texttt{AnalyzeConflict} and determines that $v_4$ causes the conflict ($v_4$ is the only variable assigned so far).  From the assignment $\overline{v_4}@1$, \tool{} learns a "backjumping" clause $v_4$, i.e., $v_4$ must be $T$. \tool{} now backtracks to $dl$ $0$ and erases all assignments decided \emph{after} this level. Thus, $v_4$ is now unassigned and the constraint  $x_1 + x_2 - 1 \le 0$ is also removed.

In \emph{iteration 3}, \texttt{BCP} determines that the learned clause is also a unit clause $v_4$ and infers $v_4@0$. In \texttt{Deduction}, we now have the new constraint $x_1 + x_2 - 1 > 0$ due to $v_4 \mapsto T$ (i.e., $x_4=\texttt{on}$).  With the new constraint, \tool{} 
approximates the output upper bound $x_5 \le  0.5$, which means $x_5>0$ might be satisfiable.
Also, \tool{} computes new bounds $0.5 \le x_3 \le 2.5$ and $0 < x_4 \le 2.0$, and deduces that $x_3$ must be positive because its lower bound is 0.5  Thus, \tool{} has a new assignment $v_3@0$ ($dl$ stays unchanged due to the implication). Note that this process of inferring new assignments from the T-solver is referred to theory propagation in DPLL(T).

In \emph{iteration 4}, \texttt{BCP} has no effects because we have no new unit clauses.  In \texttt{Deduction}, \tool{} determines that the current set of constraints, which contains the new constraint $-0.5x_1+0.5x_2+1 > 0$ (due to $v_3 \mapsto T$), is \emph{infeasible}. Thus, \tool{} enters \texttt{AnalyzeConflict} and determines that $v_4$, which was set at $dl=0$ (by \texttt{BCP} in iteration 3), causes the conflict. 
\tool{} then learns a clause $\overline{v_4}$ (the conflict occurs due to the assignment $\{v_3 \mapsto T; v_4 \mapsto T\}$, but $v_3$ was implied and thus making $v_4$ the conflict).
However, because $v_4$ was assigned at decision level 0, \tool{} can no longer backtrack and thus sets $dl=-1$ and returns \texttt{unsat}. 



This \texttt{unsat} result shows that the DNN has the property because we cannot find a counterexample violating it, i.e., no inputs $x_1 \in[-1,1] ,x_2\in [-2,2]$ that results in $x_5 > 0$.

Note that this example is too simple to illustrate the use of \emph{restart}, which is described in~\S\ref{sec:restart-tree} and \ref{sec:restart} and crucial for more complicated and nontrivial problems.


\subsection{The search tree of \tool{}}\label{sec:restart-tree}

\begin{figure}[t]
    \begin{minipage}[b]{\textwidth}
    \centering
        \begin{minipage}[t]{0.495\textwidth}
            \centering  
            \includegraphics[width=\linewidth]{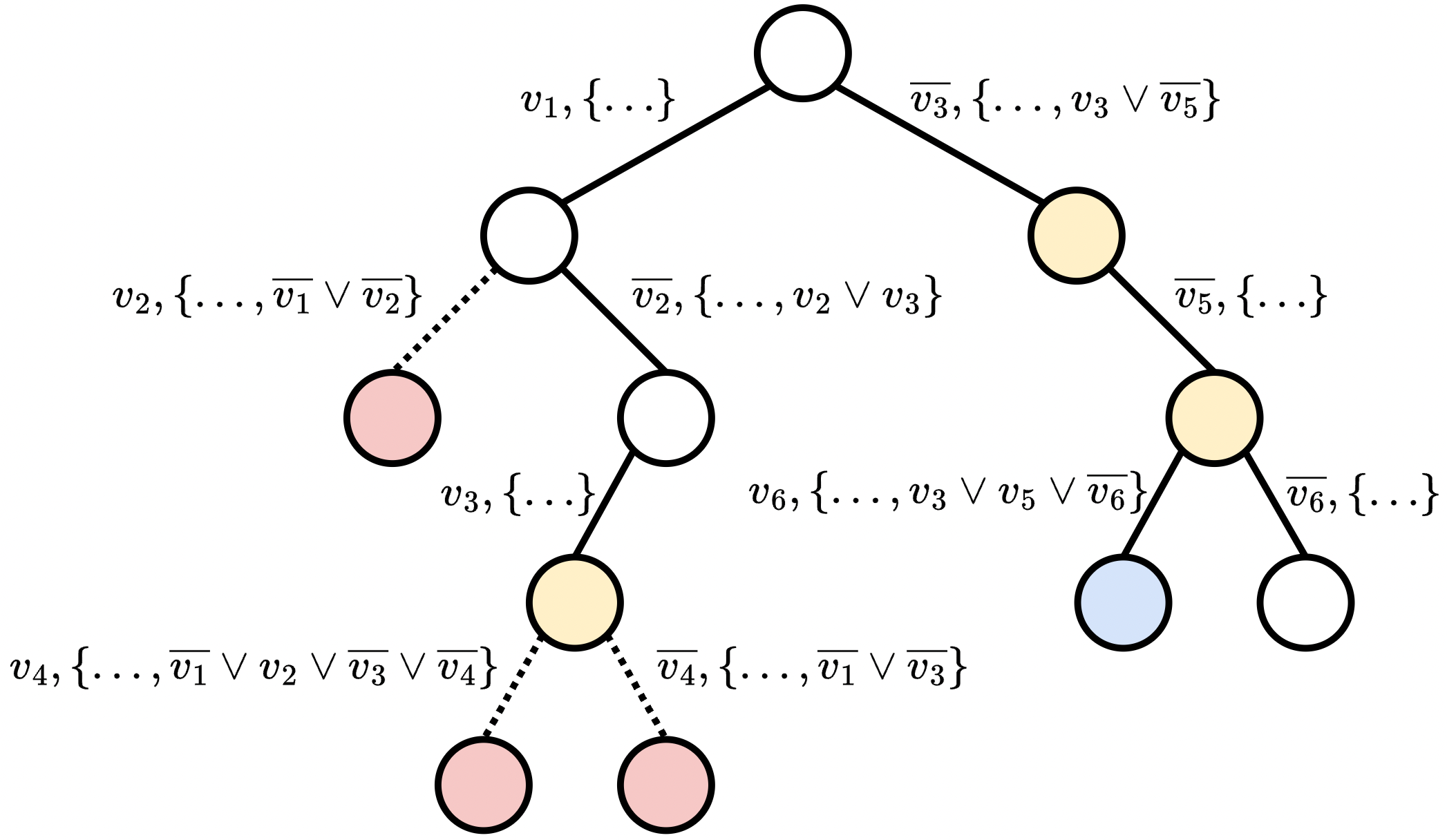}
            \caption*{(a) \tool{}}
        \end{minipage}
        \hfill
        \begin{minipage}[t]{0.44\textwidth}
            \centering
            \includegraphics[width=\linewidth]{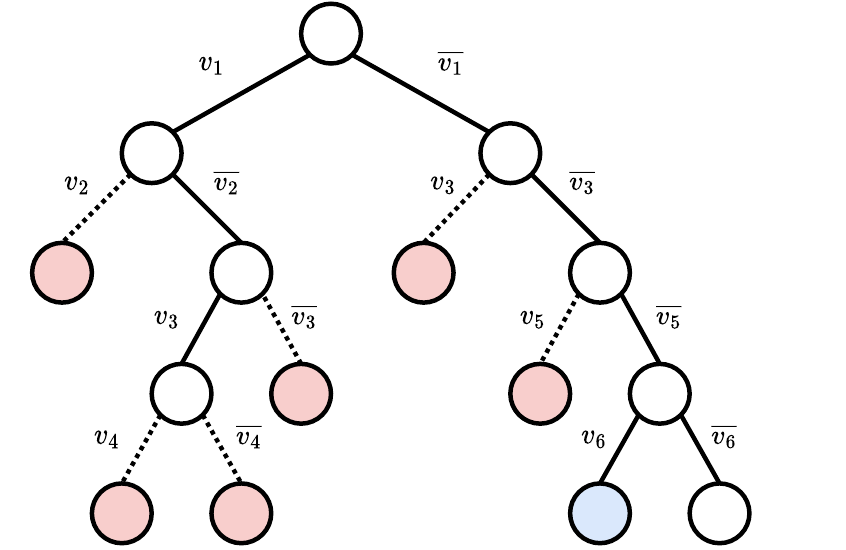}
            \caption*{(b) Others}
        \end{minipage}
        \vspace{-0.1in}
        \caption{Search tree explored by \tool{} (a) and other verifiers (b) during a verification run.
        The notation $\{...\}$ indicates learned clauses; red is  infeasibility; white is feasibility; yellow is BCP application; and blue is current consideration.
        The search tree of \tool{} is smaller than the tree of the other techniques because \tool{} was able to prune various branches, e.g., through BCPs (e.g., ${v_3}$ and $\overline{v_5}$) and non-chronological backtracks (e.g., $\overline{v_3}$).}
        \label{fig:tree}
    \end{minipage}
\end{figure}


As mentioned in \S\ref{sec:nnverif}, ReLU-based DNN verification is NP-complete, and for difficult problem instances DNN verification tools often have to exhaustively search a very large space, making scalability a main concern for modern DNN verification. 


Fig.~\ref{fig:tree} shows the difference between \tool{} and another DNN verification tool (e.g., using the popular Branch-and-Bound (BaB) approach) in how they navigate the search space.  We assume both tools employ similar abstraction and neuron splitting.
Fig.~\ref{fig:tree}b shows that the other tool performs splitting to explore different parts of the tree (e.g., splitting $v_1$ and explore the branches with $v_1=T$ and $v_1=F$ and so on). Note that the other tool needs to consider the tree shown regardless if it runs sequentially or in parallel. 

In contrast, \tool{} has a smaller search space shown in Fig.~\ref{fig:tree}a.  
\tool{} follows the path $v_1$, $v_2$ and then $\overline{v_2}$ (just like the tool on the right).
However, because of the learned clause $v_2\lor v_3$, \tool{} performs a BCP step that sets $v_3$ (and therefore prunes the branch with $\overline{v_3}$ that needs to be considered in the other tree). 
Then \tool{} splits $v_4$, and like the other tool, determines infeasibility for both branches. Now \tool{}'s conflict analysis determines from learned clauses that it needs to backtrack to $v_3$ (yellow node) instead of $v_1$.  Without learned clauses and non-chronological backtracking, \tool{} would backtrack to decision $v_1$ and continues with the $\overline{v_1}$ branch, just like the other tool in Fig.~\ref{fig:tree}b. 

Thus, \tool{} was able to generate non-chronological backtracks and use BCP to prune various parts of the search tree.  In contrast, the other tool would have to move through the exponential search space to eventually reach the same result as \tool{}.

\section{The \tool{} Approach}\label{sec:alg}

\SetKwInOut{Input}{input}
\SetKwInOut{Output}{output}
\SetKw{Break}{break}
\SetKw{Continue}{continue}
\SetKwFunction{Backtrack}{Backtrack}
\SetKwFunction{Decide}{Decide}
\SetKwFunction{BCP}{BCP}
\SetKwFunction{Deduction}{Deduction}
\SetKwFunction{AnalyzeConflict}{AnalyzeConflict}
\SetKwFunction{BooleanAbstraction}{BooleanAbstraction}
\SetKwFunction{AddClause}{AddClause}
\SetKwFunction{isTotal}{isTotal}
\SetKwFunction{randomattack}{RandomAttack}
\SetKwFunction{pgd}{PGDAttack}

\SetKwFunction{DPLLT}{DPLLT}
\SetKwFunction{isValid}{isValid}
\SetKwFunction{LPSolver}{LPSolver}
\SetKwFunction{Solve}{Solve}
\SetKwFunction{FindLayerNodes}{FindLayerNodes}
\SetKwFunction{TightenInputBounds}{TightenInputBounds}
\SetKwFunction{Abstract}{Abstract}
\SetKwFunction{Check}{Check}
\SetKwFunction{Decide}{Decide}
\SetKwFunction{Imply}{Imply}
\SetKwFunction{Lower}{LowerBound}
\SetKwFunction{Upper}{UpperBound}
\SetKwFunction{GetInputBounds}{GetInputBounds}
\SetKwFunction{GetInputs}{GetInputs}
\SetKwFunction{GetNumInputs}{GetNumInputs}
\SetKwFunction{CurrentConflictClause}{CurrentConflictClause}
\SetKwFunction{StopCriterion}{StopCriterion}
\SetKwFunction{LastAssignedLiteral}{LastLiteral}
\SetKwFunction{LiteralToVariable}{LiteralToVariable}
\SetKwFunction{Antecedent}{Antecedent}
\SetKwFunction{BinRes}{BinRes}
\SetKwFunction{BacktrackLevel}{BacktrackLevel}
\SetKwFunction{AddClause}{AddClause}
\SetKwFunction{ActivationStatus}{ActivationStatus}

\SetKwData{implicationgraph}{igraph}
\SetKwData{literal}{lit}
\SetKwData{variable}{var}
\SetKwData{antecedent}{ante}
\SetKwData{conflict}{conflict}
\SetKwData{none}{none}
\SetKwData{layerid}{lid}
\SetKwData{hiddenbounds}{hidden\_bounds}
\SetKwData{layerbounds}{lbounds}
\SetKwData{inputs}{inputs}
\SetKwData{inputbounds}{input\_bounds}
\SetKwData{outputbounds}{output\_bounds}
\SetKwData{infeasible}{INFEASIBLE}
\SetKwData{unreachable}{UNREACHABLE}
\SetKwData{maxinputs}{MAX\_NUM\_INPUT}
\SetKwData{assignment}{$\sigma$}
\SetKwData{network}{$\alpha$}
\SetKwData{dl}{dl}
\SetKwData{lpmodel}{solver}
\SetKwData{clauses}{clauses}
\SetKwData{conflict}{conflict}
\SetKwData{clause}{clause}
\SetKwData{igraph}{igraph}
\SetKwData{cex}{cex}
\SetKwData{sat}{sat}
\SetKwData{unsat}{unsat}

\SetKwData{true}{true}
\SetKwData{false}{false}
\DontPrintSemicolon




  
        
        
    
      
    
  

\SetKwFunction{Restart}{Restart}

\SetKwData{issat}{is\_sat}
\SetKwData{conflictclause}{conflict\_clause}
\SetKwData{isconflict}{is\_conflict}

\begin{algorithm}[t]
    \small
    \DontPrintSemicolon
    
    \Input{DNN $\alpha$, property $\phi_{in} \Rightarrow \phi_{out}$}
    \Output{\texttt{unsat} if the property is valid and \texttt{sat} otherwise}
    \BlankLine
    
    $\clauses \leftarrow \BooleanAbstraction(\alpha)$\;\label{line:Booleanabstraction}
  
    \While{\true}{
        $\sigma \leftarrow \emptyset$ \tcp{initial assignment}\label{line:varsa}
        $\dl \leftarrow 0$ \tcp{initial decision level}
        $\igraph \leftarrow \emptyset$ \tcp{initial implication graph}\label{line:varsb}
        \While{\true}{\label{line:dpllstart}
            $\isconflict \leftarrow \true$ \;
        
            \If{$\BCP(\clauses, \sigma, \dl, \igraph)$}{\label{line:bcp}
                \If{$\Deduction(\sigma, \dl, \alpha, \phi_{in}, \phi_{out})$}{ \label{line:deduction}
                    $\issat, v_i \leftarrow \Decide(\alpha, \phi_{in}, \phi_{out}, \dl, \sigma)$ \tcp{decision heuristic} \label{line:decide} 
                    \lIf(\tcp*[h]{total assignment}){\issat}{
                        \Return{$\sat$} \label{line:returnsat} 
                    }
                    $\sigma \leftarrow \sigma \land v_i$\;
                    $\dl \leftarrow \dl + 1$\;
                    $\isconflict \leftarrow \false$ \tcp{mark as no conflict}
                }
            }
            \If{\isconflict}{
                \lIf(\tcp*[h]{conflict at decision level 0}){$\dl \equiv 0$}{ 
                    \Return{\unsat}  \label{line:unsat}
                }
                $\clause \leftarrow \AnalyzeConflict(\igraph)$\;
                $\dl \leftarrow \Backtrack(\sigma, \clause)$ \;\label{line:backtrack}
                $\clauses \leftarrow \clauses \cup \{\clause\}$ \tcp{learn conflict clauses} \label{line:learn}
            }
            
            \lIf(\tcp*[h]{restart heuristic}){$\Restart{}$}{\label{line:restart}
                \Break
            }
        }\label{line:dpllend}
    }
    \caption{The  \tool{}  DPLL(T) algorithm.}\label{fig:alg}
  
\end{algorithm}

Fig.~\ref{fig:alg} shows the \tool{} algorithm, which takes as input the formula $\alpha$ representing the ReLU-based DNN $N$ and the formulae $\phi_{in}\Rightarrow \phi_{out}$ representing the property $\phi$ to be proved.
Internally, \tool{} checks the satisfiability of the formula
\begin{equation}\label{eq:prob}
  \alpha \land \phi_{in} \land \overline{\phi_{out}}.
\end{equation}
\tool{} returns \texttt{unsat} if the formula unsatisfiable, indicating  that $\phi$ is a valid property of $N$, and \texttt{sat} if it is satisfiable, indicating the $N$ is not a valid property of $\phi$.

\tool{} uses a  DPLL(T)-based algorithm to check unsatisfiability.
First, the input formula in Eq.~\ref{eq:prob} is abstracted to a propositional formula
with variables encoding neuron activation status (\texttt{BooleanAbstraction}).
Next, \tool{} assign values to Boolean variables (\texttt{Decide}) and checks for conflicts the assignment has with the real-valued constraints of the DNN and the property of interest (\texttt{BCP} and \texttt{Deduction}). 
If conflicts arise, \tool{} determines the assignment decisions causing the conflicts (\texttt{AnalyzeConflict}), backtracks to erase such decisions (\texttt{Backtrack}), and learns clauses to avoid those decisions in the future. 
\tool{} repeats these decisions and checking steps until it finds a total or full assignment for all Boolean variables, in which it returns \texttt{sat}, or until it no longer can backtrack, in which it returns \texttt{unsat}.
Note that \tool{} also resets its search (if it thinks that it is stuck in a local optima) and tries different decision orderings to enable better clause learning and avoid similar ``bad'' decisions in the previous runs.
We describe these steps in more detail below.

\subsection{Boolean Abstraction}
\texttt{BooleanAbstraction} (Fig.~\ref{fig:alg} line~\ref{line:Booleanabstraction}) encodes the DNN verification problem into a Boolean constraint to be solved by DPLL.  This step creates Boolean variables to represent the \emph{activation status} of hidden neurons in the DNN. Observe that when evaluating the DNN on any concrete input, the value of each hidden neuron \emph{before} applying ReLU is either $>0$ (the neuron is \emph{active} and the input is passed through to the output) or $\le 0$ (the neuron is \emph{inactive} because the output is 0).
This allows partial assignments to these variables to represent neuron activation patterns within the DNN.

From the given network, \tool{} first creates Boolean variables representing the activation status of neurons. Next, \tool{} forms a set of initial clauses ensuring that each status variable is either \texttt{T} or \texttt{F}, indicating that each neuron is either active or inactive, respectively.
For example, for the DNN in Fig.~\ref{fig:dnn}, \tool{} creates two status variables $v_3,v_4$ for neurons $x_3,x_4$, respectively, and two initial clauses $v_3\lor \overline{v_3}$ and $v_4 \lor \overline{v_4}$. The assignment $\{x_3=T, x_4=F \}$ creates the constraint $0.5x_1-0.5x_2-1>0 \land x_1 + x_2 -2 \le 0$.

\subsection{DPLL}\label{sec:dpll}

After \texttt{BooleanAbstraction}, \tool{} iteratively searches for an assignment satisfying the status clauses (Fig.~\ref{fig:alg}, lines~\ref{line:dpllstart}--~\ref{line:dpllend}).
\tool{} combines  DPLL components (e.g., \texttt{Decide}, \texttt{BCP}, \texttt{AnalyzeConflict}, \texttt{Backtrack} and \texttt{Restart}) to assign truth values with a theory solver (\S\ref{sec:deduction}), consisting of abstraction and linear programming solving, to check the feasibility of the constraints implied by the assignment with respect to the network and property of interest.

\tool{} maintains several variables (Fig.~\ref{fig:alg}, lines~\ref{line:Booleanabstraction}--~\ref{line:varsb}). These include $\clauses$, a set of \emph{clauses} consisting of the initial activation clauses and learned clauses;   $\sigma$, a \emph{truth assignment} mapping status variables to truth values; $igraph$, an \emph{implication graph} used for analyzing conflicts; and  $dl$, a non-zero \emph{decision level} used for assignment and backtracking.



\subsubsection{Decide}\label{sec:decide}

From the current assignment, \texttt{Decide} (Fig.~\ref{fig:alg}, line~\ref{line:decide}) uses a heuristic to choose an unassigned variable and assigns it a random truth value at the current decision level. 
\tool{} applies the Filtered Smart Branching (FSB) heuristic~\cite{bunel2018unified,de2021improved}. For each unassigned variable, FSB assumes that it has been decided (i.e., the corresponding neuron has been split) and computes a fast approximation of the lower and upperbounds of the network output variables.  FSB then prioritizes unassigned variables with the best differences among the bounds that would help make the input formula unsatisfiable (which helps prove the property of interest). 
Note that if the current assignment is full, i.e., all variables have assigned values, \texttt{Decide} returns \texttt{False} (from which \tool{} returns \texttt{sat}).

\subsubsection{Boolean Constraint Propagation (BCP)}\label{sec:bcp}

From the current assignment and clauses, \texttt{BCP} (Fig.~\ref{fig:alg}, line~\ref{line:bcp}) detects \emph{unit clauses}\footnote{A unit clause is a clause that has a single unassigned literal.} and infers values for variables in these clauses.
For example, after the decision $a\mapsto F$, \texttt{BCP} determines that the clause $a\vee b$ becomes unit, and infers that $b \mapsto T$.
Moreover, each assignment due to \texttt{BCP} is associated with the current decision level because instead of being ``guessed'' by \texttt{Decide} the chosen value is logically implied by other assignments.
Moreover, because each {BCP} implication might cause other clauses to become unit, \texttt{BCP} is applied repeatedly until it can no longer find unit clauses.
\textsc{BCP} returns \texttt{False} if it obtains contradictory implications (e.g., one BCP application infers $a \mapsto F$ while another infers $a \mapsto T$), and returns \texttt{True} otherwise.

\paragraph{Implication Graph} \texttt{BCP} uses an \emph{implication graph}~\cite{barrett2013decision} to represent the current assignment and the reason for each BCP implication. In this graph, a node represents the assignment and an edge $i \xrightarrow{c} j$ means that \texttt{BCP} infers the assignment represented in node $j$ due to the unit clause $c$ caused by the assignment represented by node $i$.
The implication graph is used by both BCP, which iteratively constructs the graph on each BCP application and uses it to determine conflict, and \texttt{AnalyzeConflict} (\S\ref{sec:analyze-conflicts}), which analyzes the conflict in the graph to learn clauses.

\begin{figure}
    \begin{minipage}[c]{0.17\textwidth}
        \centering
        \small
        \begin{equation*}
            \begin{aligned}
                c_1 &= (\overline{v_1} \lor v_2) \\
                c_2 &= (\overline{v_1} \lor v_3 \lor v_5) \\
                c_3 &= (\overline{v_2} \lor v_4) \\
                c_4 &= (\overline{v_3} \lor \overline{v_4}) 
            \end{aligned}      
        \end{equation*}
        \caption*{(a)}
    \end{minipage}
    \begin{minipage}[c]{0.40\textwidth}
        \centering
        \includegraphics[width=1\linewidth]{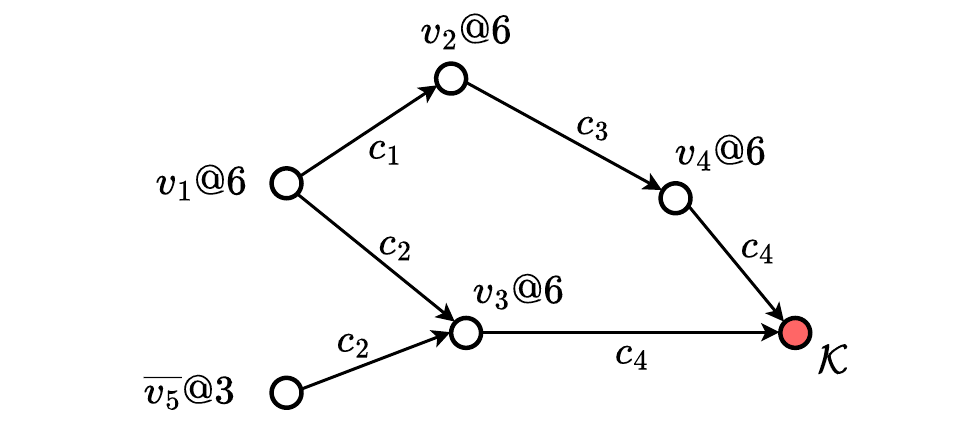}
        \caption*{(b)}
    \end{minipage}
    \begin{minipage}[c]{0.33\textwidth}
        \centering
        \small
        \begin{tabular}{ccccc}
            name & cl & lit & var & ante \\
            \midrule
            $c_4$ &$\overline{v_3} \lor \overline{v_4}$   & $\overline{v_3}$ &$v_3$   &     $c_2$\\
            & $\overline{v_4} \lor \overline{v_1} \lor v_5$  & $\overline{v_4}$   &     $v_4$  &      $c_3$\\
            &$\overline{v_1} \lor v_5 \lor \overline{v_2}$   & $\overline{v_2}$   &     $v_2$  &      $c_1$\\
            $c_5$      & $\overline{v_1} \lor v_5$
        \end{tabular}
        \caption*{(c)}        
    \end{minipage}
    \caption{(a) A set of clauses, (b) an implication graph, and (c) learning a new clause.\label{fig:igraph}}
\end{figure}

\paragraph{Example}  Assume we have the clauses in Fig.~\ref{fig:igraph}(a), the assignments $\overline{v_5}@3$ and $v_1@6$ (represented in the graph in Fig.~\ref{fig:igraph}(b) by nodes  $\overline{v_5}@3$ and $v_1@6$, respectively), and are currently at decision level $dl$ 6.
Because of assignment $v_1@6$, \texttt{BCP} infers $v_2@6$ from the unit clause $c_1$ and captures that implication with edge $v_1@6 \xrightarrow{c_1} v_2@6$.
Next, because of assignment $v_2@6$, \texttt{BCP} infers $v_4@6$ from the unit clause $c_3$ as shown by edge $v_2@6 \xrightarrow{c_3} v_4@6$.

Similarly, \texttt{BCP} creates edges $v_1@6 \xrightarrow{c_2} v_3@6$ and $\overline{v_5}@ \xrightarrow{c_2} v_3@6$ to capture  the inference $v_3@6$ from the unit clause $c_2$ due to assignments $\overline{v_5}@3$ and $v_1@6$.
Now, \texttt{BCP} detects a conflict because clause $c_4=\overline{v_3} \lor \overline{v_4}$ cannot be satisfied with the assignments $v_4@6$ and $v_3@6$ (i.e., both $v_3$ and $v_4$ are $T$)  and creates two edges to the (red) node $\kappa$: $v_4@6 \xrightarrow{c_4} \kappa$ and $v_3@6 \xrightarrow{c_4} \kappa$ to capture this conflict. 

Note that in this example \texttt{BCP} has the implication order $v_2,v_4,v_3$ (and then reaches a conflict). In the current implementation, \tool{} makes an arbitrary decision and thus could have a different order, e.g., $v_3, v_4,v_2$.

\subsubsection{Conflict Analysis}\label{sec:analyze-conflicts}

Given an implication graph with a conflict such as the one in Fig.~\ref{fig:igraph}(b), \texttt{AnalyzeConflict} learns a new \emph{clause} to avoid past decisions causing the conflict.
The algorithm traverses the implication graph backward, starting from the conflicting node $\kappa$, while constructing a new clause through a series of resolution steps.
\texttt{AnalyzeConflict} aims to obtain an \emph{asserting} clause, which is a clause that will force an immediate BCP implication after backtracking.

\begin{wrapfigure}{r}{0.50\textwidth}
\begin{minipage}{\linewidth}
\vspace{-0.2in}
\begin{algorithm}[H]
    \small
    \Input{implication graph \implicationgraph}
    \Output{\clause}
    \BlankLine
    $\clause \gets \CurrentConflictClause(\implicationgraph)$\;\label{line:conflict}
    \While{$\neg \StopCriterion(\clause)$}{\label{line:loopstart} 
      $\literal \gets \LastAssignedLiteral(\implicationgraph, \clause)$\;\label{line:lit}
      $\variable \gets \LiteralToVariable(\literal)$\;\label{line:var}      
      $\antecedent \gets \Antecedent(\implicationgraph,\literal)$\;\label{line:ante}
      $\clause \gets \BinRes(\clause, \antecedent, \variable)$\;\label{line:resolve}
    }\label{line:loopend}
    \Return{$\clause$}
    \caption{\textsc{AnalyzeConflict}}\label{alg:conflict}
  \end{algorithm}
  \end{minipage}
\end{wrapfigure}
\texttt{AnalyzeConflict}, shown in Fig.~\ref{alg:conflict}, first extracts the conflicting clause $cl$ (line~\ref{line:conflict}), represented by the edges connecting to the conflicting node $\kappa$ in the implication graph.
Next, the algorithm refines this clause to achieve an asserting clause (lines~\ref{line:loopstart}--~\ref{line:loopend}).
It obtains the literal $lit$ that was assigned last in $cl$ (line~\ref{line:lit}), the variable $var$ associated with $lit$ (line~\ref{line:var}), and the antecedent clause $ante$ of that $var$ (line~\ref{line:ante}), which contains $\overline{lit}$ as the only satisfied literal in the clause. Now, \texttt{AnalyzeConflict} resolves $cl$ and $ante$ to eliminate literals involving $var$ (line~\ref{line:resolve}). The result of the resolution is a clause, which is then refined in the next iteration.

\paragraph{Resolution.} We use the standard \emph{binary resolution rule} to learn a new clause implied by two (\emph{resolving}) clauses $a_1 \lor ... \lor a_n \lor \beta$ and $b_1 \lor ... \lor b_m \lor \overline{\beta}$ containing complementary literals involving the (\emph{resolution}) variable $\beta$:
\begin{equation} \label{eq:binary-resolution}
    \frac{(a_1 \lor ... \lor a_n \lor \beta) \quad \quad (b_1 \lor ... \lor b_m \lor \overline{\beta})}{(a_1 \lor ... \lor a_n \lor b_1 \lor ... \lor b_m)} \quad  (\textsc{Binary-Resolution})
\end{equation}
The resulting (\emph{resolvant}) clause $a_1 \lor ... \lor a_n \lor b_1 \lor ... \lor b_m$ contains all the literals that do not have complements $\beta$ and $\neg{\beta}$.



\paragraph{Example} Fig.~\ref{fig:igraph}(c) demonstrates \texttt{AnalyzeConflict} using the example in \S\ref{sec:bcp} with the BCP implication order $v_2,v_4,v_3$ and the conflicting clause $cl$ (connecting to node $\kappa$ in the graph in Fig.~\ref{fig:igraph}(b)) $c_4=\overline{v_3} \lor \overline{v_4}$. From $c_4$, we determine the last assigned literal is $lit=\overline{v_3}$, which contains the variable $var=v_3$, and the antecedent clause containing $v_3$ is $c_2=\overline{v_1} \lor v_3 \lor v_5$ (from the implication graph in Fig.~\ref{fig:igraph}(b), we determine that assignments $v_1@6$ and $\overline{v_5}@3$ cause the BCP implication $v_3@6$ due to clause $c_2$). Now we resolve the two clauses $cl$ and $c_2$ using the resolution variable $v_3$ to obtain the clause $\overline{v_4} \lor \overline{v_1} \lor v_5$.
Next, from the new clause, we obtain $lit=\overline{v_4}, var=v_4, ante=c_3$ and apply resolution to get the clause $\overline{v_1} \vee v_5 \lor \overline{v_2}$.
Similarly, from this clause, we obtain $lit=\overline{v_2}, var=v_2, ante=c_1$ and apply resolution to obtain the clause $v_1 \lor v_5$.
                 

At this point, \texttt{AnalyzeConflict} determines that this is an asserting clause, which would force an immediate BCP implication after backtracking. As will be shown in \S\ref{sec:backtrack}, \tool{} will backtrack to level 3 and erases all assignments after this level (so the assignment $\overline{v_5}@3$ is not erased, but assignments after level 3 are erased).  Then, \texttt{BCP} will find that $c_5$ is a unit clause because $\overline{v_5}@3$ and infers $\overline{v_1}$.
Once obtaining the asserting clause, \texttt{AnalyzeConflict} stops the search, and \tool{} adds  $v_1\lor v_5$ as the new clause $c_5$ to the set of existing four clauses.

The process of learning clauses allows \tool{} to learn from its past mistakes.
While such clauses are logically implied by the formula in Eq.~\ref{eq:prob} and therefore do not change the result, they help prune the search space and allow DPLL and therefore \tool{} to scale. For example, after learning the clause $c_5$, together with assignment $v_5@3$, we immediately infer $v_1 \mapsto F$ through BCP instead of having to guess through \texttt{Decide}.

\subsubsection{Backtrack} \label{sec:backtrack} 
From the clause returned by \texttt{AnalyzeConflict}, \texttt{Backtrack}  (Fig.~\ref{fig:alg}, line~\ref{line:backtrack}) computes a backtracking level and erases all decisions and implications made after that level.
If the clause is \emph{unary} (containing just a single literal), then we backtrack to level 0. 

Currently, \tool{} uses the standard \emph{conflict-drive backtracking} strategy~\cite{barrett2013decision}, which sets the backtracking level to the \emph{second most recent} decision level in the clause.
Intuitively, by backtracking to the second most recent level, which means erasing assignments made \emph{after} that level, this strategy encourages trying new assignments for more recently decided variables.


\paragraph{Example} From the clause $c_5=\overline{v_1} \lor v_5$ learned in \texttt{AnalyzeConflict}, we backtrack to decision level 3, the second most recent decision level in the clause (because assignments $v_1@6$  and $\overline{v_5}@3$ were decided at levels 6 and 3, respectively). Next, we erase all assignments from decision level 4 onward (i.e., the assignments to $v_1,v_2,v_3,v_4$ as shown in the implication graph in Fig.~\ref{fig:igraph}). This thus makes these more recently assigned variables (after decision level 3) available for new assignments (in fact, as shown by the example in \S\ref{sec:bcp}, \texttt{BCP} will immediately infer $v_1=T$ by noticing that $c_5$ is now a unit clause).






\subsubsection{Restart}\label{sec:restart}

As with any stochastic algorithm, \tool{} can perform poorly if it gets into a subspace of the search that does not quickly lead to a solution, e.g., due to choosing a bad sequence of neurons to split~\cite{bunel2018unified,de2021improved}.  
This problem, which has been recognized in early SAT solving, motivates the introduction of restarting the search~\cite{gomes1998boosting} to avoid being stuck in such a \emph{local optima}.



\tool{} uses a simple restart heuristic (Fig.~\ref{fig:alg}, line~\ref{line:restart}) that triggers a restart when either the number of processed assignments (nodes) exceeds a pre-defined number (e.g., 300 nodes) or the current runtime exceeds a pre-defined threshold (e.g., 50 seconds).
After a restart, \tool{} avoids using the same decision order of previous runs (i.e., it would use a different sequence of neuron splittings). It also resets all internal information (e.g., decisions and implication graph) except the learned conflict clauses, which are kept and reused as these are \textit{facts} about the given constraint system.
This allows a restarted search to quickly prune parts of the  space of assignments.


We found the combination of clause learning and restarts effective for DNN verification. In particular, while restart resets information  it keeps learned clauses, which  are \emph{facts} implied by the problem, and therefore enables quicker BCP applications and non-chronological backtracking (e.g., as illustrated in Fig.~\ref{fig:tree}).

It is worth noting that while it is possible to add a restart to existing DNN verification approaches. This is unlikely to help, because these techniques do not learn conflict clauses and therefore restart will just randomize order but carry no information forward to prune the search space.



\subsection{Deduction (Theory Solving)}\label{sec:deduction}

\begin{algorithm}[t]
    \small

    \Input{ DNN $\network$, input property $\phi_{in}$, output property $\phi_{out}$, decision level $dl$ and current assignment $\assignment$}
    \Output{{\false} if infeasibility occurs, {\true} otherwise}

    \BlankLine

    $\lpmodel \leftarrow \LPSolver(\assignment, \network \land \phi_{in} \land \overline{\phi_{out}})$\;\label{line:lpsolver}
    \lIf{$\Solve(\lpmodel) \equiv \infeasible$}{\Return {\false}}\label{line:solve}
    \lIf(\tcp*[h]{orig prob (Eq.~\ref{eq:prob}) is satisfiable}){$\isTotal(\assignment)$}{\Return {\true} }\label{line:full}


    $\inputbounds \leftarrow \TightenInputBounds(\lpmodel, \phi_{in})$\;\label{line:absstart}
    $\outputbounds, \hiddenbounds \leftarrow \Abstract(\network, \assignment, \inputbounds$)\;\label{line:abstraction}

    \lIf{$\Check(\outputbounds, \overline{\phi_{out}}) \equiv \infeasible$}{\Return {\false}}\label{line:checkfeasibility}

    \For{$v \in \hiddenbounds$} {\label{line:infera}
      $x \leftarrow \ActivationStatus(v)$\;
      \lIf{$x \in \sigma \lor \neg{x} \in \sigma$}{\Continue}
      \lIf{$\Lower(v) > 0$}{$\assignment \leftarrow \assignment \cup x@dl$}
        \lElseIf{$\Upper(v) \le 0$}{$\assignment \leftarrow \assignment \cup \overline{x}@dl$}\label{line:inferb}
    }

    \Return {\true}\;\label{line:absend}

    \caption{\textsc{Deduction}}\label{alg:deduction}
\end{algorithm}

\texttt{Deduction} (Fig.~\ref{fig:alg}, line~\ref{line:deduction}) is the theory or T-solver, i.e., the T in DPLL(T). The main purpose of the T-solver is to check the feasibility of the constraints represented by the current propositional variable assignment; as shown in the formalization in \S\ref{sec:formalization} this amounts to just \emph{linear equation} solving for verifying piecewise linear DNNs. However, \tool{} is able to leverage specific information from the DNN problem, including input and output properties, for more aggressive feasibility checking.  Specifically, \texttt{Deduction} has three tasks: (i) checking feasibility using linear programming (LP) solving, (i) further checking feasibility with input tightening and abstraction, and (iii) inferring literals that are unassigned and are implied by the abstracted constraint.


Fig.~\ref{alg:deduction} describes \texttt{Deduction}, which returns \texttt{False} if infeasibility occurs and  \texttt{True} otherwise.
First, it creates a linear constraint system from the input assignment $\sigma$ and $\alpha \land \phi_{in} \land \overline{\phi_{out}}$, i.e., the formula in Eq.~\ref{eq:prob} representing the original problem  (line~\ref{line:lpsolver}).
The key idea is that we can remove ReLU activation for hidden neurons whose activation status have been decided. 
For constraints in $\alpha$ associated with variables that are not in the $\sigma$,  we ignore them and just consider the cutting planes introduced by the partial assignment.  
For example, for the assignment $v_3\mapsto T, v_4 \mapsto F$, the non-linear ReLU constraints $x_3=ReLU(-0.5x_1+0.5x_2+1)$ and  $x_4=ReLU(x_1+x_2-1)$ for the DNN in Fig.~\ref{fig:dnn} become linear constraints $x_3=-0.5x_1+0.5x_2$ and $x_4=0$, respectively.

Next, an LP solver checks the feasibility of the linear constraints (line~\ref{line:solve}).
If the solver returns infeasible,  \texttt{Deduction} returns \texttt{False} so that \tool{} can analyze the assignment and backtrack. 
If the constraints are feasible, then there are two cases to handle. First, if the assignment is total (i.e., all variables are assigned), then that means that the original problem is satisfiable (line~\ref{line:full}) and \tool{} returns \texttt{sat}.

\paragraph{ReLU Abstraction.} Second, if the assignment is not total then \texttt{Deduction} applies abstraction to check satisfiability (lines~\ref{line:absstart}--\ref{line:checkfeasibility}).
Specifically, we over-approximate  ReLU computations to obtain the upper and lower bounds of the output values and check if the output properties are feasible with respect to these bounds. For example, the output $x_5 > 0$ is \emph{not} feasible if the upperbound is $x_5 \le 0$ and \emph{might be} feasible if the upperbound is $x_5 \le 0.5$ (``might be'' because this is an upper-bound). If abstraction results in infeasibility, then \texttt{Deduction} returns \texttt{False} for \tool{} to analyze the current assignment (line~\ref{line:checkfeasibility}).

\begin{figure}
    \begin{minipage}[b]{1\linewidth}
        \centering
        \begin{minipage}[c]{0.24\linewidth}
            \includegraphics[width=\linewidth]{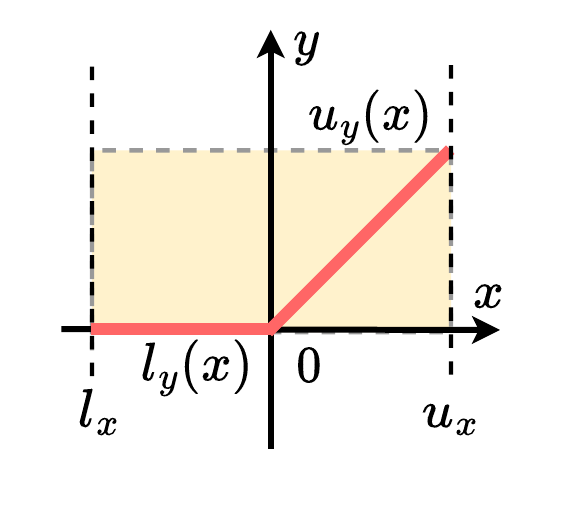}
            \vspace*{-10mm}
            \caption*{(a) interval}
        \end{minipage}
        \begin{minipage}[c]{0.24\linewidth}
            \includegraphics[width=\linewidth]{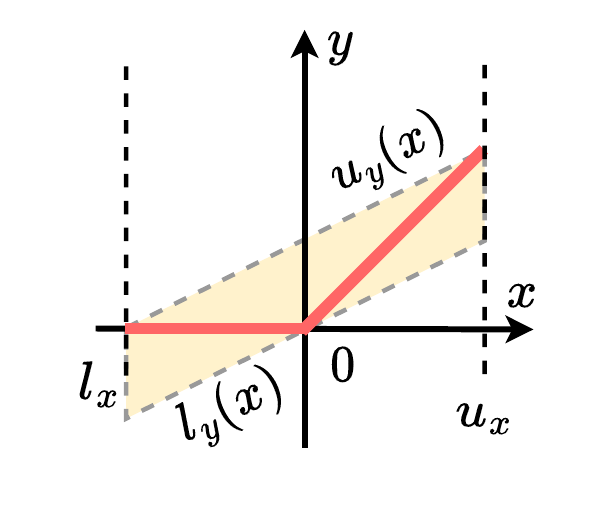}
            \vspace*{-10mm}
            \caption*{(b) zonotope}
        \end{minipage}
        \begin{minipage}[c]{0.24\linewidth}
            \includegraphics[width=\linewidth]{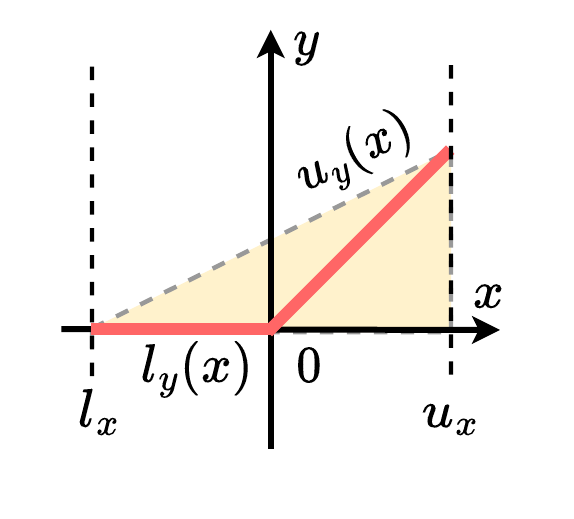}
            \vspace*{-10mm}
            \caption*{(c) DeepPoly}
        \end{minipage}
        \begin{minipage}[c]{0.24\linewidth}
            \includegraphics[width=\linewidth]{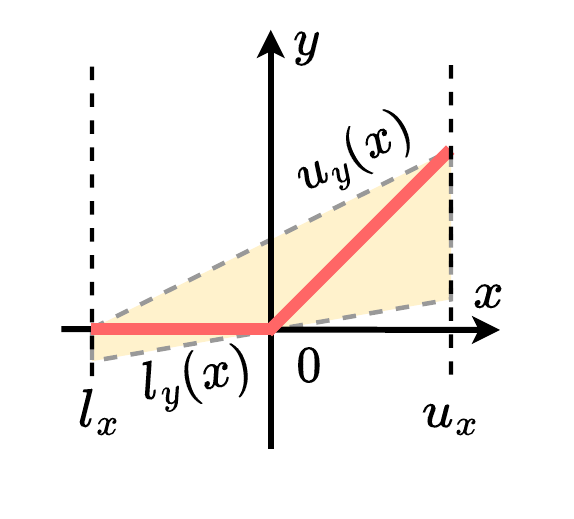}
            \vspace*{-10mm}
            \caption*{(d) LiRPA}
        \end{minipage}
        \vspace*{-3mm}
        \caption{Abstractions for ReLU: (a) interval, (b) zonotope, and (c-d) polytopes. Notice that ReLU is a non-convex region (red line) while all abstractions are convex regions. Note that (c) and (d) are both polytopes.}\label{fig:abs}
    \end{minipage}
\end{figure}

\tool{} uses abstraction to approximate the lower and upper bounds of hidden and output neurons.
Fig.~\ref{fig:abs} compares the (a) interval~\cite{wang2018formal}, (b) zonotope~\cite{singh2018fast}, and (c, d) polytope~\cite{xu2020fast,singh2019abstract,wang2021beta} abstraction domains to compute the lower $l_y(x)$ and upper $u_y(x)$ bounds of a ReLU computation $y=\texttt{ReLU(x)}$ (non-convex red line).
\tool{} can employ any existing abstract domains, though currently it adopts the \emph{LiRPA} polytope (Fig.~\ref{fig:abs}d)~\cite{xu2020fast,wang2021beta} because it has a good trade-off between precision and efficiency.


\paragraph{Inference} If abstraction results in feasible constraints, \texttt{Deduction} next attempts to infer implied literals (lines~\ref{line:infera}--~\ref{line:inferb}). To obtain the bounds of the output neurons, abstraction also needs to compute the bounds of hidden neurons, including those with undecided activation status (i.e., not yet in $\sigma$). 
This allows us to assign the activation variable of a hidden neuron the value
\texttt{True} if the lowerbound of that neuron is greater than 0 (the neuron is active) and
\texttt{False} otherwise.
Since each literal is considered, this would be considered exhaustive theory propagation.  Whereas the literature~\cite{nieuwenhuis2006solving,kroening2016decision} suggests that this is an inefficient strategy, we find that it does not incur significant overhead (average overhead is about 4\% and median is 2\% with outliners being large CIFAR2020 networks described in \S\ref{sec:expsettings}).  


\paragraph{Example} For the illustrative example in \S\ref{sec:unsat}, in iteration 3, the current assignment $\sigma$ is  $\{v_4=1\}$, corresponding to a constraint $x_1 + x_2 - 1 > 0$. With the new constraint, we optimize the input bounds and compute the new bounds for hidden neurons $0.5 \le x_3 \le 2.5$, $0 < x_4 \le 2.0$ and output neuron  $x_5 \le 0.5$ (and use this to determine that the postcondition $x_5 > 0$ might be feasible). We also infer $v_3=1$ because of the positive lower bound $0.5 \le x_3$.

\subsection{Optimizations}\label{sec:optimizations}
Like some other verifiers~\cite{katz2019marabou,katz2022reluplex,bak2021nnenum}, 
\tool{} implements \emph{input splitting} to quickly deal with \emph{small} verification problems, such as ACAS Xu discussed in \S\ref{sec:expsettings}. 
This technique divides the original verification problem into subproblems, each checking whether the DNN produces the desired output from a smaller input region and returns \texttt{unsat} if all subproblems are verified and \texttt{sat} if a counterexample is found in any subproblem.

Moreover, like other DNN verifiers~\cite{ferrari2022complete,zhang2022general}, \tool{} tool implements a fast-path optimization that attempts to disprove or falsify the property before running DPLL(T).
\tool{} uses two \emph{adversarial attack} algorithms to find counterexamples to falsify properties.
First, we try a randomized attack approach~\cite{das2021fast}, which is a derivative-free sampling-based optimization~\cite{yu2016derivative}, to generate a potential counterexample.
If this approach fails, we then use a gradient-based approach~\cite{madry2017towards} to create another potential counterexample. 
If either attack algorithm gives a valid counterexample, \tool{} returns \texttt{sat}, indicating that property is invalid. If both algorithms cannot find a valid counterexample or they exceed a predefined timeout, \tool{} continues with its DPLL(T) search.

\section{Implementation and Experimental Settings}\label{sec:expsettings}

\paragraph{Implementation} \tool{} is written in Python 
and uses PyTorch~\cite{paszke2019pytorch} for matrix multiplications and Gurobi~\cite{gurobi} for linear constraint solving. We use the LiRPA abstraction library~\cite{xu2020fast,wang2021beta} for bounds approximation and tightening and adapt the randomized~\cite{das2021fast} and Projected Gradient Descent (PGD)~\cite{madry2017towards} adversarial attack techniques for falsification (\S\ref{sec:optimizations}).



Currently, \tool{} supports feedforward (FNN), convolutional (CNN), and Residual Learning Architecture (ResNet) neural networks that use ReLU.
\tool{} automatically preprocesses these networks into Boolean variables and linear constraints representing the computation graph of the networks for DPLL(T). This preprocessing step is relatively standard and is used by various tools (e.g., \crown{}, \mnbab{}). Moreover, \tool{} supports the specification formats ONNX~\cite{onnx2} for neural networks and VNN-LIB~\cite{vnnlib} for properties.
These formats are standard and supported by major DNN verification tools. 

\begin{table*}
    \footnotesize
    \caption{Benchmark instances. U: \texttt{unsat}, S: \texttt{sat}, ?: \texttt{unknown}.}\label{tab:benchmarks}
    \vspace*{-3mm}
    
    \begin{tabular}{c|cccc|cc}
        \toprule
        \multirow{2}{*}{\textbf{Benchmarks}} &\multicolumn{2}{c}{\textbf{Networks}} &  \multicolumn{2}{c|}{\textbf{Per Network}} &\multicolumn{2}{c}{\textbf{Tasks}} \\
        & Type & Networks & Neurons & Parameters & Properties & Instances (U/S/?) \\
        \midrule
        
        ACAS Xu & FNN & 45 & 300 & 13305 & 10 & 139/47/0 \\
        \midrule
    
        MNISTFC & FNN & 3 &  0.5--1.5K & 269--532K & 30 & 56/23/11\\
        \midrule
        
        CIFAR2020 & FNN+CNN & 3 &  17--62K & 2.1--2.5M & 203 & 149/43/11 \\
        \midrule
        
        RESNET\_A/B & CNN+ResNet & 2 & 11K & 354K & 144 & 49/23/72 \\
        \midrule
        
        CIFAR\_GDVB & FNN+CNN & 42 & 9--49K & 0.08--58M & 39 & 60/0/0 \\
                                
        \midrule
        \textbf{Total} & & \textbf{95}  & & & \textbf{426} & \textbf{453/136}/94 \\
        
        \bottomrule
        \end{tabular}
\end{table*}

\paragraph{Benchmarks} We evaluate \tool{} using four standard ReLU-based benchmarks obtained from \vnncomp{} -- shown in the first 4 rows of Tab.~\ref{tab:benchmarks} -- and a new benchmark \textbf{CIFAR\_GDVB} that we describe below.
In total these benchmarks consist of 95 networks, spanning multiple layer types and architectures, and 426 safety/robustness properties. A problem instance pairs a property with a network.  
Across the benchmark 453 problem instances are known to be \texttt{unsat} (U)
and 136 are known to be \texttt{sat} (S).   For the remaining 94 problem instances no
verifier in our study, or in \vnncomp{}, was able to solve the problem (?).

\textbf{ACAS Xu} consists of 45 FNNs to issue turn advisories to aircrafts to avoid collisions. Each FNN has 5 inputs (speed, distance, etc). 
We use all 10 safety properties as specified in~\cite{katz2017reluplex} and \vnncomp{}, where properties 1--4 are used on 45 networks and properties 5--10 are used on a single network.
\textbf{MNISTFC} consists of 3 FNNs for handwritten digit recognition and 30 robustness properties. 
Each FNN has 1x28x28 inputs representing a handwritten image.
\textbf{CIFAR2020} has 3 CNNs for objects detection and 203 robustness properties (each CNN has a set of different properties). Each network uses 3x32x32 RGB input images.
For \textbf{RESNET\_A/B}, each benchmark has only one network with the same architecture and 72 robustness properties. Each network uses 3x32x32 RGB input images. 

The organizers of VNN-COMP have observed that many of these benchmarks are \textit{easy} in the sense that they can be proven or disproven in just a few seconds~\cite{muller2022third}. 
Such benchmarks encourage verifier developers to include \textit{fast-path} optimizations that perform, e.g., adversarial attacks to falsify properties or single-pass overapproximations to prove properties.  Benchmarks which are \textit{hard} provide an opportunity to assess how verification algorithms confront the combinatorial complexity of DNN verification.
We developed a hard benchmark by leveraging a systematic DNN verification problem generator GDVB~\cite{xu2020systematic}.  Briefly, GDVB takes a seed neural network as input and systematically varies a number of architectural parameters, e.g., number of layers, and neurons per layer, to produce a new benchmark.  In this experiment, we begin with a single CIFAR network with 3 convolutional layers and 1 fully-connected layer, then generate 42 different DNNs that cover combinations of parameter variations and 39 local robustness properties with varying radii and center points.   
Problems that could be solved by \crown{} and \mnbab{} within 20 seconds were removed from the benchmark as ``too easy''.
This resulted in a new benchmark \textbf{CIFAR\_GDVB} containing 60 verification problems that not only are more computationally challenging than benchmarks used in other studies, e.g.,~\cite{muller2022third}, but also exhibit significant architectural diversity.

Tab.~\ref{tab:benchmarks} provides more details. 
Column \textbf{Instances (U/S/?)} shows the number of verification instances and U/S/? indicate the number of instances that are \texttt{unsat} (valid property), \texttt{sat} (invalid property), and \texttt{unknown} (cannot be solved by any tools and we also do not know if it is \texttt{sat} or \texttt{unsat}), e.g., 
CIFAR2020 has 203 instances (149 \texttt{unsat}, 43 \texttt{sat}, and 11 \texttt{unknown}). 
The \textbf{Per Network} column gives the sizes of individual networks:  \textbf{Neurons} are the numbers of hidden neurons and \textbf{Parameters} are the numbers of weights and biases. 
For example, each FNN in ACAS Xu has 5 inputs, 6 hidden layers (each with 50 neurons), 5 outputs, and thus has 300 neurons ($6 \times 50$) and 13305 parameters ($5\times 50\times 50 + 2\times 50 \times 5 + 6 \times 50 + 5$).

\paragraph{Verification Tools.} 
We analyzed the top-3 best performing DNN verifiers in the 2021 and 2022 instances
of VNN-COMP to select a set of state-of-the-art verifiers that includes:
\textbf{\crown{}}~\cite{wang2021beta,zhang2022general} ranked first in both competitions, and
\textbf{\mnbab{}}~\cite{ferrari2022complete} ranked second in 2022 and its predecessor
\textbf{\eran{}}~\cite{muller2021scaling,singh2019beyond,singh2019abstract} ranked third in 2021. 
\verinet{}~\cite{henriksen2020efficient} ranked second in 2021 and third in 2022, but it requires a license
for a commercial constraint solver that we were unable to acquire.
We replaced \verinet{} with \textbf{\nnenum{}}~\cite{bak2020improved,bak2021nnenum} which ranked fourth in the 2022 competition.  All of these verifiers employ abstraction and a form of branch-and-bound reasoning, but they do not perform any form of CDCL or the optimizations it enables.

We also selected the versions of 
\textbf{\marabou{}}~\cite{katz2019marabou,katz2022reluplex} from both 2021 and 2022 since they 
represent the best-performing constraint-based DNN verifiers -- making them comparable
to \tool{}.  We note that the versions from each competition exhibit different performance, which
is why we include them both.

\S\ref{sec:related} provides additional detail on the algorithmic techniques implemented in these verifiers.

\paragraph{Hardware and Setup.}
Our experiments were run on Linux with AMD Threadripper 64-core 4.2GHZ CPU, 128GB RAM, and NVIDIA GeForce 4090 GPU with 24 GB VRAM. 
All tools use multiprocessing in some way (e.g., external tools/libraries including Gurobi, LiRPA, and Pytorch are multi-thread).  
\crown{}, and \mnbab{} leverage GPU processing for abstraction. The LiRPA library adopted by \tool{} uses the GPU for large benchmarks. 


To maximize the performance of the DNN verifiers in comparisons and to promote replicability, we leverage the benchmarks and
installation scripts available from VNN-COMP\footnote{\url{https://github.com/ChristopherBrix/vnncomp2022_benchmarks}}.
These scripts were tailored by the developers of each verifier to optimize
performance on each benchmark.
The VNN-COMP setting used varying runtimes for each problem instance ranging from 120 seconds to more than 12 minutes.  We experimented with timeouts on our machine and settled on 1800 seconds for instances of CIFAR\_GDVB and 900 seconds of other instances.  These timeouts allowed verifiers to achieve their good scoring performance reported in VNN-COMP'22. For each benchmark instance we run three times and obtain the median results.

\section{Results}\label{sec:eval}
We evaluate \tool{} to answer the following research questions:
\begin{description}
\item[RQ1 (\S\ref{sec:ablation})]{How do clause-learning and restart impact \tool{} performance?}
\item[RQ2 (\S\ref{sec:vnncomp})]{How does \tool{} compare to state-of-the-art DNN verifiers?}
\item[RQ3 (\S\ref{sec:comparison-marabou})]{How does \tool{} compare to DPLL(T)-based DNN verifiers?}
\end{description}

We note that in our experiments all tools provide \emph{correct results}. If a tool was able to solve an instance, then it solves it correctly, i.e., no tool returned \texttt{sat} for \texttt{unsat} instances and vice versa. 
\subsection{RQ1: Clause-learning and Restart Ablation Study}\label{sec:ablation}
\tool{}'s CDCL and restart functionality offer potential benefits in mitigating the exponential cost of verification.  We apply two treatments to explore those benefits:
``Full''  corresponds to the full algorithm in Fig.~\ref{fig:alg}; and
``No Restart''  corresponds to algorithm in Fig.~\ref{fig:alg} without \texttt{Restart} (Line~\ref{line:restart}).

\begin{figure}[t]
    \hfill
    \begin{subfigure}[b]{0.24\textwidth}
        \centering
        \footnotesize
        \resizebox{\linewidth}{!}{%
        \begin{tabular}{c|c|c}
            \toprule
            \textbf{Setting} & \textbf{\#Solved} &\textbf{Avg. Time} \\
            \midrule
            No Restart       & 38    & 1269.4 \\
            Full             & 58    & 1100.1 \\
            \bottomrule    
        \end{tabular}
        }
        \vspace*{4mm}
        \caption{}\label{tab:cdcl_vs_dpll}
     \end{subfigure}
     \hfill
     \begin{subfigure}[b]{0.34\textwidth}
         \centering
        \includegraphics[width=\linewidth]{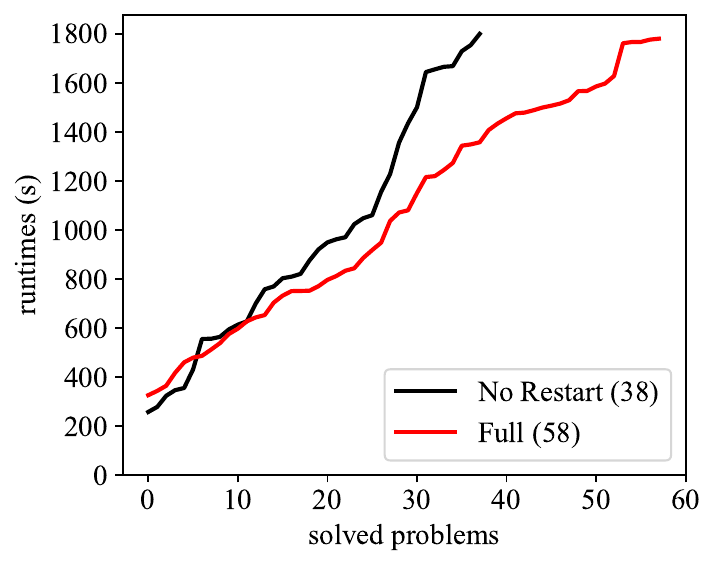}
        \caption{}\label{fig:cdcl_vs_dpll}
     \end{subfigure}
     \hfill
     \begin{subfigure}[b]{0.37\textwidth}
        \centering
        \includegraphics[width=\linewidth]{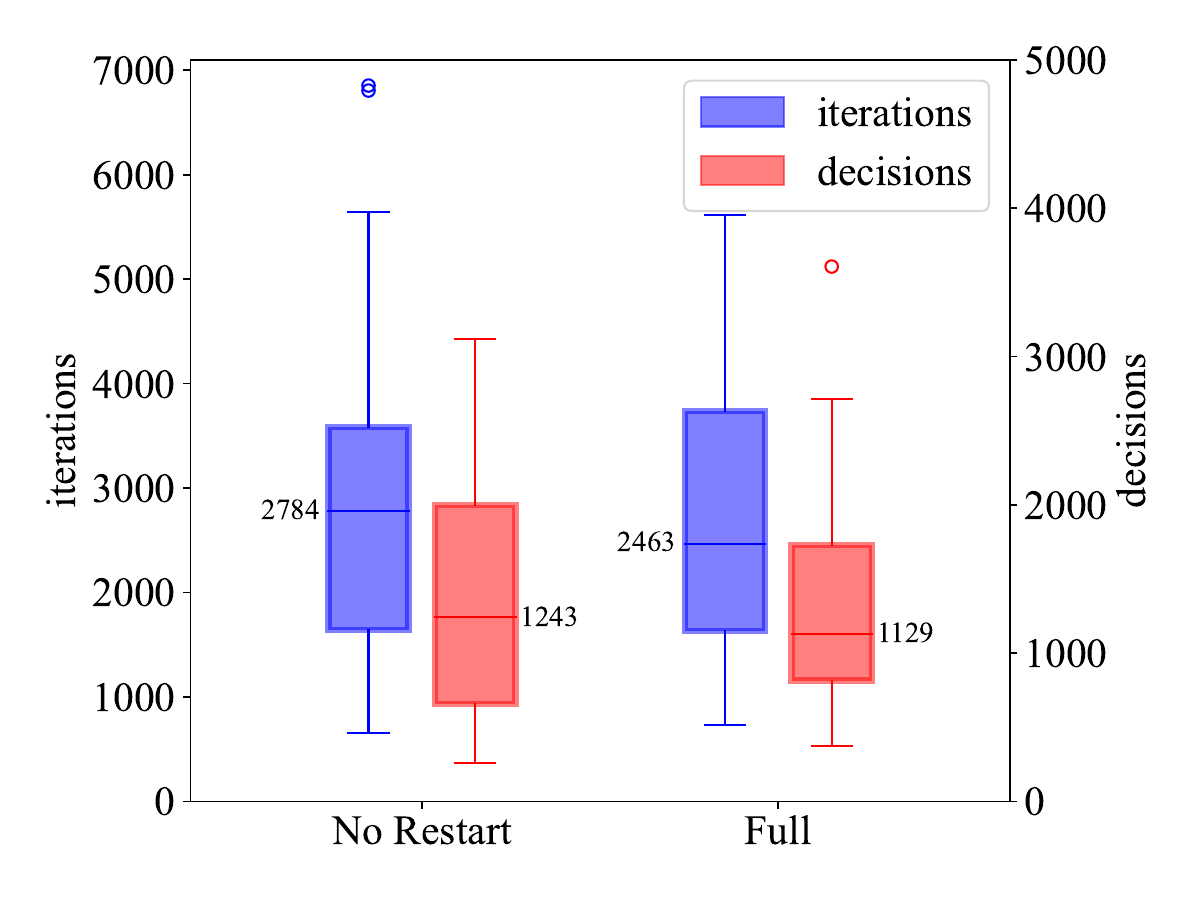}
        \caption{}\label{fig:iteration_decision}
     \end{subfigure}
     \vspace{-2mm}
    \caption{Performance of \tool{} with ``Full'' CDCL settings and with ``No Restart'' on CIFAR\_GDVB benchmark using three different metrics: (a) Problems solved and solve time (s); (b) Sorted solved problems; and (c) Comparing counts of iterations and decisions.}\label{fig:rq1}
\end{figure}

We use the 60 challenging CIFAR\_GDVB instances in this study
since they force the verifier to explore the exponentially sized space of variable assignments; eliminating the potential for fast-path optimizations.
Our primary metrics are the number of verification 
problems solved and the time to solve them.

Fig.~\ref{fig:rq1}(a) presents data on \tool{} with different treatments (Settings) in the table.
Fig.~\ref{fig:rq1}(b) shows the problems solved within the 1800-second timeout for each technique sorted by runtime from fastest to slowest; problems that timeout are not shown on the plot.

These data clearly show the benefit of CDCL with restart.
Compared to no restart, 20 additional problems can be verified
which represents a 53\% increase.
We note that performing restarts causes the search process
to begin again and potentially performs redundant analysis, but
the search after a restart carries forward the learned clauses
which serve to prune subsequent search.
Fig.~\ref{fig:rq1}(b) illustrates the overhead incurred by
restart in the fastest 10 problems -- before the ``Full'' and ``No Restart'' curves diverge.
We note that performing a restart without the clauses learned through CDCL amounts to rerunning the verifier with a different random seed to vary search order.   While this could be achieved
with other DNN verifiers its benefit for verification would be limited.

To further understand the benefits of CDCL with and without restart, we collected internal data from \tool{} to record
the number of iterations of in Fig.~\ref{fig:alg} and the number of decisions computed (Line~\ref{line:decide}) on average across the benchmark.  Note that for the ``No Restart'' treatment the outer loop executes a single time so the number of iterations are just for the inner loop.
Fig.~\ref{fig:iteration_decision} plots the median and quartiles for iterations needed on the left axis and for decisions made on the right axis with box plots.
We annotate the median values next to the box plots.
These data show that restarts lead to a reduction
in both median number of iterations and decisions, 11\% and 9\%, respectively.  

For these experiments, we allowed only 3 restarts that were triggered when either
300 branches were explored or 50 seconds had elapsed, so at most 4 iterations
of the outer loop in in Fig.~\ref{fig:alg} were executed.  Despite these restarts
the number of iterations of the inner loop was reduced indicating
that later restart phases were able to accelerate through the search space using
learned clauses.  The data for decisions tell a consistent story since learned
clauses will allow BCP to prune branches at decision points in later restart phases.

\subsection{RQ2: Comparison with State-of-the-art Verifiers}\label{sec:vnncomp}
To compare verifiers, 
we adopt the rules in \vnncomp{} to score and rank tools.  
For each verification problem instance, a tool scores 10 points if it correctly verifies an instance, 1 point if it correctly falsifies an instance, and 0 points if it cannot solve (e.g., timeouts, has errors, or returns \texttt{unknown}), -150 points if it gives incorrect results (this penalty did not apply in the scope of our study).
We note that \vnncomp{} assigns different scores for falsification: 1 point if the tool found a counterexample using an external adversarial attack technique, and 10 points if the tool found a counterexample using its core search algorithm. The tools we compared to did not report how they
falsified problems, so we give a single point for a false result regardless of how it was obtained.
We note that \tool{} exhibited the best falsification performance so it is likely disadvantaged by this scoring approach.

\begin{table}[]
     \caption{A \textbf{Verifier}'s rank (\textbf{\#}) is based on its VNN-COMP score (\textbf{S}) on a {benchmark}. For each benchmark, the number of problems verified (\textbf{V}) and falsified (\textbf{F}) are shown.}\label{tab:score}

\resizebox{\textwidth}{!}{
\begin{tabular}{c|cccc|cccc|cccc|cccc|cccc|cccc}
\toprule
\multirow{2}{*}{\textbf{Verifier}} &
  \multicolumn{4}{c|}{\textbf{ACAS Xu}} &
  \multicolumn{4}{c|}{\textbf{MNISTFC}} &
  \multicolumn{4}{c|}{\textbf{CIFAR2020}} &
  \multicolumn{4}{c|}{\textbf{RESNET\_A/B}} &
  \multicolumn{4}{c|}{\textbf{CIFAR\_GDVB}} &
  \multicolumn{4}{c}{\textbf{Overall}} \\

 & \textbf{\#} & \textbf{S} & \textbf{V} & \textbf{F} &
   \textbf{\#} & \textbf{S} & \textbf{V} & \textbf{F} &
   \textbf{\#} & \textbf{S} & \textbf{V} & \textbf{F} &
   \textbf{\#} & \textbf{S} & \textbf{V} & \textbf{F} &
   \textbf{\#} & \textbf{S} & \textbf{V} & \textbf{F} &
   \textbf{\#} & \textbf{S} & \textbf{V} & \textbf{F} \\
\midrule

\crown{} &
2 & 1436 & \textbf{139} & 46 & 
\textbf{1} & \textbf{582} & \textbf{56} & 22 & 
\textbf{1} & \textbf{1522} & \textbf{148} & 42 & 
\textbf{1} & \textbf{513} & \textbf{49} & \textbf{23} & 
\textbf{1} & \textbf{600} & \textbf{60} & 0 & 
\textbf{1} & \textbf{4653} & \textbf{452} & 133 \\ 
\midrule

\tool{} &
4 & 1417 & 137 & \textbf{47} & 
5 & 363 & 34 & \textbf{23} & 
3 & 1483 & 144 & \textbf{43} & 
2 & 403 & 38 & \textbf{23} & 
2 & 580 & 58 & 0 & 
2 & 4246 & 411 & \textbf{136} \\ 
\midrule

\mnbab{} &
5 & 1097 & 105 & \textbf{47} & 
3 & 370 & 36 & 10 & 
2 & 1486 & 145 & 36 & 
3 & 363 & 34 & \textbf{23} & 
3 & 470 & 47 & 0 & 
3 & 3786 & 367 & 116 \\ 
\midrule

\nnenum{} &
\textbf{1} & \textbf{1437} & \textbf{139} & \textbf{47} & 
2 & 403 & 39 & 13 & 
4 & 518 & 50 & 18 & 
- & - & - & - & 
- & - & - & - & 
4 & 2358 & 228 & 78 \\ 
\midrule

\marabouold{} &
3 & 1426 & 138 & 46 & 
4 & 370 & 35 & 20 & 
- & - & - & - & 
- & - & - & - & 
- & - & - & - & 
5 & 1796 & 173 & 66 \\ 
\midrule

\marabounew{} &
6 & 1015 & 97 & 45 & 
6 & 308 & 29 & 18 & 
- & - & - & - & 
- & - & - & - & 
- & - & - & - & 
6 & 1323 & 126 & 63 \\ 

\bottomrule

\end{tabular}
}
\end{table}

Tab.~\ref{tab:score} shows the results of \tool{} and the top-performing VNN-COMP verifiers:
\crown{}, \mnbab{}, and \nnenum{}, and two versions of \marabou{}.
We report the rank (\textbf{\#}) and score (\textbf{S}) of each tool using the VNN-COMP rules for each benchmark as well as the overall rank.
Tools that do not work on a benchmark are not shown under that benchmark (e.g., \marabou{} reports errors for all CIFAR2020 problems).
The last two columns break down the number of problems each verifier was able to verify (\textbf{V}) or falsify (\textbf{F}).

Across these benchmarks \tool{} ranks second to \crown{}, which was the top performer in \vnncomp{} and thus the state-of-the-art.   It trails \crown{} in the number of problems verified, though it can falsify more problems than any other verifier across the benchmarks.

Both \marabou{} and \nnenum{} outperform \tool{} on the \textbf{MNISTFC} and \textbf{ACAS Xu},
but we observe that these are small DNNs.   On the larger
DNNs in the \textbf{CIFAR2020}, \textbf{RESNET\_A/B} and \textbf{CIFAR\_GDVB} benchmarks, which
have orders of magnitude more neurons, \tool{} significantly outperforms those techniques.

While ranking second, \tool{} solves 95\% of the problems in the large network benchmarks that are solved by \crown{}.  Moreover, on the most challenging benchmark, \textbf{CIFAR\_GDVB}, 
\tool{} solves 2 fewer problems than \crown{}.
We expect that further optimization of \tool{} will help close that gap and note that \crown{} has been under development for over 4 years and is highly-optimized from years of VNN-COMP participation.
In addition, \crown{}'s developers tuned 10 parameters, on average,
to optimize its performance for each individual benchmark.
In contrast, we did not tune any parameters for \tool{} which suggests that its performance on large models may generalize better in practice and that further improvement could come from parameter tuning.

\subsection{RQ3: Comparison with DPLL(T)-based DNN Verifiers}\label{sec:comparison-marabou}
The state-of-the-art in DPLL(T)-based DNN verification is \marabou{}.
It improves on \reluplex{} by incorporating abstraction and deduction techniques, and 
has been entered in \vnncomp{} in recent years.  This makes it a reasonable point of
comparison for \tool{} especially in understanding the benefit of the addition of
CDCL on the scalability of DNN verification.

Overall both versions of \marabou{} ranked poorly, but it did outperform \tool{} on 
small DNNs.  Consider the ACAS Xu networks which are small in two ways: they have very few neurons (300) and they only have 5 input dimensions. 
\marabou{} employs multiple optimizations to target small scale networks.
For example, a variant of the Split and Conquer algorithm~\cite{wu2020parallelization} 
subdivides the input space to generate separate verification problems.
Partitioning a 5 dimensional input space is one thing, but the number of partitions
grows exponentially with input dimension and this approach is not cost
effective for the larger networks in our study.

\marabou{} could not scale to any of the larger CIFAR or RESNET problems, so a direct comparison with \tool{} is not possible.
Instead, we observe that \tool{} performed well on these problems -- ranking better than it did on the smaller problems.

We conjecture that this is because problems of this scale give ample time for
clause learning and CDCL to significantly prune the search performed by DPLL(T).
Evidence for this can be observed in data on the learned clauses recorded during
runs of \tool{} on \texttt{unsat} problems.  Since \tool{}'s propositional
encodings have a number of variables proportional to the number of neurons ($n$)
in the network the effect of a learned clause of
size $c$ is that it has the potential to block a space of assignments of
size $2^{n - c}$.  In other words, as problems grow the reduction through CDCL grows
combinatorially.  In the largest problem  in the benchmarks, with $n = 62464$ we see clauses on average 
of size $c = 16$ which allows BCP to prune an enormous space of assignments -- of size $2^{62448}$.   
The ability of \tool{} to scale well beyond other DPLL(T) approaches to DNN verification demonstrates the benefit of CDCL.

\section{Related Work}\label{sec:related}
The literature on DNN verification is rich and is steadily growing (cf.~\cite{urban2021review,liu2021algorithms}). Here we summarize well-known techniques with tool implementations.


\textbf{Constraint-based} approaches such as 
\texttt{DLV}~\cite{huang2017safety},
\texttt{Planet}~\cite{ehlers2017formal}, and  
\reluplex{}~\cite{katz2017reluplex} and its successor  \marabou{}~\cite{katz2019marabou,katz2022reluplex} transform DNN verification into a constraint problem, solvable using an SMT (Planet, DLV) or DPLL-based search with a customized simplex and MILP solver (\reluplex{}, \marabou{}) solvers.
\textbf{Abstraction-based} techniques and tools
such as  \texttt{AI\(^2\)}~\cite{gehr2018ai2}, 
\texttt{ERAN}~\cite{muller2021scaling,singh2019abstract,singh2018fast}
(\texttt{DeepZ}~\cite{singh2018fast},
\texttt{RefineZono}~\cite{singh2018boosting},
\texttt{DeepPoly}~\cite{singh2019abstract}, 
\texttt{K-ReLU}~\cite{singh2019beyond}), 
\mnbab{}~\cite{ferrari2022complete}), 
\reluval{}~\cite{wang2018formal}, \neurify{}~\cite{wang2018efficient}, \verinet~\cite{henriksen2020efficient}, \nnv{}~\cite{tran2021robustness}, \nnenum{}~\cite{bak2021nnenum,bak2020improved}, \texttt{CROWN}~\cite{zhang2018efficient}, \crown{}~\cite{wang2021beta}, use abstract domains such as interval (\reluval{}/\neurify{}), zonotope (\texttt{DeepZ}, \nnenum{}), polytope (\texttt{DeepPoly}), starset/imagestar (\nnv{}, \nnenum{}) to scale verification.
\texttt{OVAL}~\cite{ovalbab} and \dnnv{}~\cite{shriver2021dnnv} are frameworks employing various existing DNN verification tools.
Our \tool{}, which is most related to \marabou{}, is a DPLL(T) approach that integrates clause learning and abstraction in theory solving.


Well-known \textbf{abstract domains} for DNN verification include interval, zonotope, polytope, and starset/imagestar. Several top verifiers such as \mnbab{} and \nnenum{} use multiple abstract domains (e.g., \mnbab{} uses deeppoly and lirpa, \nnenum{} adopts deeppoly, zonotope and imagestar. The work in~\cite{goubault2021static} uses the general max-plus abstraction~\cite{heidergott2006max} to represent the non-convex behavior of ReLU. 
\tool{} currently uses polytope in its theory solver though it can also use other abstract domains.

Modern SAT solvers benefit from effective \textbf{heuristics}, e.g., VSIDS and DLIS strategies for decision (branching), random restart~\cite{moskewicz2001chaff} and shortening~\cite{chinneck1991locating} or deleting clauses~\cite{moskewicz2001chaff} for memory efficiency and avoiding local maxima caused by greedy strategies. 
Similarly, modern DNN verifiers such as \nnenum{}, \crown{}, and \marabou{} include many \textbf{optimizations} to improve performance, e.g., Branch-and-Bound~\cite{bunel2020branch} and Split-and-Conquer~\cite{katz2019marabou,katz2022reluplex,wu2020parallelization} for parallelization, and various optimizations for abstraction refinement~\cite{singh2018boosting,bak2021nnenum}) and bound tightening~\cite{bak2021nnenum,katz2019marabou,wang2021beta}.
\tool{} has many opportunities for improvements such as new decision heuristics and parallel DPLL(T) search algorithms (\S\ref{sec:conclusion}).

\section{Conclusion and Future Work}\label{sec:conclusion}

We introduce \tool{}, a DPLL(T) approach and prototype tool for DNN verification. \tool{} includes the standard DPLL components such as clause learning, non-chronological backtracking, and restart heuristics in combination with a theory solver customized for DNN reasoning. 
We evaluate the \tool{} prototype with standard FNNs, CNNs, and Resnets, and show that \tool{} is  competitive to the state-of-the-art DNN verification tools. 

Despite its relatively unoptimized state, \tool{} already demonstrates competitive performance compared to optimized state-of-the-art DNN verification tools (\S\ref{sec:eval}). 
By adopting the DPLL(T) framework, \tool{} presents an opportunity to explore how
additional optimizations and frameworks developed for SMT can be adapted to support DNN verification.

\paragraph{Using Existing CDCL Heuristics and Optimizations} 
Currently, \tool{} incorporate standard "textbook" implementation of CDCL components, such as Filtered Smart Branching (FSB)~\cite{de2021improved} for decision heuristics. To enhance performance, we are investigating and adapting advanced SAT/SMT solver techniques, including parallel search~\cite{le2017painless,le2019modular}, alternative decision heuristics~\cite{al2022boosting,cai2022better}, and more diverse restarting strategies~\cite{guo2012heuristic,liang2018machine} for \tool{}.

\paragraph{Exploiting DNN-specific analyses} 
By having clear separate DNN-specific components, e.g., T-Solver, \tool{} can seamlessly integrate new advancements for DNN analyses and optimizations. Currently \tool{} uses polytope, but we can easily transition to alternative abstractions, such as star sets~\cite{tran2021verification}.  We are exploring other abstract refinement and bounds tightening techniques for DNNs. For example, determining the \emph{neuron stability}~\cite{xiao2018training,zhangheng2022can,chen2022linearity} (i.e., whether a neuron is active or inactive for all inputs satisfying preconditions) to eliminate ReLU computations and subsequently reduce decision choices.

\paragraph{Inheriting Formal Properties} By following a DPLL(T)-based algorithm, \tool{} inherits formal properties established for DPLL(T). In \S\ref{sec:formalization}, we adapt the transition rules and formal proofs developed for DPLL(T) approaches in~\cite{nieuwenhuis2006solving} to show that, similar to other DPLL-based SAT or SMT solvers with LRA theory, \tool{} is sound, complete, and terminates. 
    
\paragraph{Proof Generating Capabilities} DNN verification techniques (e.g., Marabou~\cite{isac2022neural}) can be unsound  and prove invalid properties--a highly undesirable behavior for formal verification tools.
The DPLL(T)-based \tool{} already uses implication graphs and resolution rules for conflict analysis and thus includes a native mechanism to produce unsat proofs, which we can extend with minimal overhead to extract proof certificates (~\cite{asin2008efficient,zhang2003validating}). 
\paragraph{Model Debugging} By being a SAT solver, we compute \emph{unsat core}~\cite{kroening2016decision}, i.e., an unsatisfiable subset of the original set of clause (or unsatisfiable activation patterns for DNN) and adapt optimizations to minimize unsat cores~\cite{zhang2003extracting}.
Unsat cores are useful for various debugging tasks, e.g., the Alloy specification analyzer~\cite{jackson2000alcoa} and the fault localization work in~\cite{zheng2021flack} use unsat cores to identify inconsistencies in a specification. 
Similarly, a user of \tool{} can use unsat cores to understand why an unexpected property is valid in a DNN or why we cannot produce a counterexample for a presumably invalid property.

\bibliographystyle{ACM-Reference-Format}
\bibliography{paper,matt}

\appendix
\section{\tool{} DPLL(T) Formalization}\label{sec:formalization}

In \S\ref{sec:alg} we describe \tool{} and its optimizations. 
Here we formalize the \tool{} DPLL(T) framework.
By abstracting away heuristics, optimizations, and implementation details, we can focus on the core \tool{} algorithm and establish its correctness and termination properties.

\tool{} can be described using the states and transition rules of the standard DPLL(T) framework described in ~\cite{nieuwenhuis2006solving} and therefore inherits the theoretical results established there.
We also highlight the differences between \tool{} and standard DPLL(T), but these differences do not affect any main results. The section aims to be self-contained, but readers who are familiar with the work in~\cite{nieuwenhuis2006solving} can quickly skim through it.

\subsection{Preliminaries} 
\paragraph{Formulae, Theory, and Satisfiability} Let $P$ be a finite set of atoms (e.g., linear constraints in our context). For an atom $p\in P$, $p$ is a positive literal and $\neg p$ is a negative literal of $P$. A \emph{clause} is a set of literals and a CNF \emph{formula} is a set of clauses.  
A model $M$ is a sequence of literals and never contains both a literal and its negation.

A literal $l$ is \emph{true} in $M$ if $l \in M$, is \emph{false} in $M$ if $\neg l \in M$, and is \emph{undefined} otherwise (i.e., $l \notin M$). $M$ is \emph{total} if every atom $p\in P$ has a literal in $M$, and is \emph{partial} otherwise.
A clause $C$ is true in $M$, written as $M \models C$, if $\exists l \in C.~ l \in M$, and is false in $M$, written as $M \models \neg C$, if $\forall l \in C.~\neg l \in M$. A CNF $F$ is true in (or satisfied by) $M$, written as $M \models F$, if all clauses of $F$ are true in $M$. In that case, $M$ is called a \emph{model} of $F$.  If F has no models then it is \emph{unsatisfiable}. 
If $F$ and $F'$ are formulae, 
then $F$ \emph{entails} $F'$, 
written as $F \models F'$, if $F'$ is true in all models of $F$.
Note that we consider literals and clauses as purely boolean variables and check satisfiability using propositional logic (we could also treat literals as syntactical items and check satisfiability using set operations, e.g., $M \models C$ is $C \cap M \neq \emptyset $).

A theory T is a set of formulas. A formula F is \emph{T-satisfiable} or \emph{T-consistent} if $F \land T$ is satisfiable. Otherwise, it is called \emph{T-unsatisfiable} or \emph{T inconsistent}. 
An assignment $M$ can be thought as a conjunction of its literals and hence as a formula. If $M$ is a T-consistent and $F$ is a formula 
such that $M \models F$, then $M$ is also a \emph{T-model} of F, written as $M \models_T F$.
If F and G are formulae, then F entails G in T , written $F \models_T G$ if $F \land \neg G$ is T-inconsistent. 
Note when checking satisfiability in the theory, i.e., $\models_T$, we use a theory solver to reason about the linear constraints represented by the literals.

\textbf{\tool{} Algorithm.}  For \tool{}, each atom $p_i$ in $P=\{p_1,p_2,\dots,p_N\}$ is the linear constraint representing activation status of neuron $i$, e.g., for the DNN example in Fig.~\ref{fig:dnn}, $p_3$ is $-0.5x_1 + 0.5x_2 + 1 > 0$, the constraint that neuron $x_3$ is active (thus $p_3$ is a positive literal and $\neg p_3$ is a negative literal).
$M$ represents the (partial) truth assignment $\sigma$, and $F$ represents the set or conjunction of clauses that \tool{} needs to satisfy. 
Adding $l$ to $M$ is the truth assignment $p \mapsto T$ if $l$ is $p$, and is the assignment $p \mapsto F$  if $l$ is $\neg p$. Moreover, the theory we consider is LRA (Linear Real Arithmetic) and our customized T-solver, described in ~\ref{sec:deduction} uses LP solving and abstraction to decide satisfiability of DNN properties.

Note that in \tool{} we use Boolean Abstraction to create variables $v_i$ to represent linear constraints capturing activation status.
Here we do not use Boolean Abstraction and capture its effects with atoms $p_i$ representing the Boolean variables $v_i$ and adding to $M$ the literals $p_i, \neg p_i$ corresponding to truth assignments $v_i\mapsto T, v_i\mapsto F$ in $\sigma$, respectively.


\subsection{Transition Rules}\label{sec:transition}

We formalize the \tool{} DPLL(T) using \emph{transition rules} that move from a state to another state of the algorithm. A \emph{state} is either a assignment $M$ and a CNF formula $F$, written as $M \parallel F$, or the special state \texttt{Fail}, which indicates that the formula is unsatisfiable.
We write $S \Longrightarrow S'$ as a transition from state $S$ to $S'$. We write $S \Longrightarrow^* S'$ to indicate any possible transition from $S$ to $S'$ (i.e., reflexive-transitive closure). 
In a state $M \parallel F, C$, we say the clause $C$ is conflicting  if  $M \models \neg C$.

\begin{table}
    \caption{Transition rules for \tool{} DPLL(T) solver.}\label{tab:rules}
    \centering    
    \resizebox{\textwidth}{!}{
    \begin{tabular}{c|ccccl}
    &\textbf{Rule} & From & & To &\textbf{Condition} \\
    \midrule{}
    \multirow{8}{*}{\rotatebox[origin=c]{90}{\textbf{Standard DPLL}}}&\textbf{Decide} & \( M~ \parallel ~ F \) & $\longrightarrow$ &  \( M~l^d~\parallel ~F \) &  
    \textbf{if}\quad \(
    \begin{cases}
        l \notin M\\
        l \text{ or } \neg l \text{ occurs in } F\\
    \end{cases}
    \)    \\\\
    &\textbf{BCP} & \( M~ \parallel ~ F,~ C \lor l \) & \( \longrightarrow \) & \( M~l ~\parallel ~ F, ~C \lor l \) & 
    \textbf{if}\quad \(
    \begin{cases}
        l \notin M\\
        M \models \neg C
    \end{cases}
    \)\\\\ 
    &\textbf{Fail} & \( M~ \parallel ~ F,~C  \) & \( \longrightarrow \) & \texttt{Fail} & 
    \textbf{if}\quad \(
    \begin{cases}
        M \text{ contains no decision literals}
        \\
        M \models \neg C
    \end{cases}
    \)\\\\
    \midrule
    \multirow{8}{*}{\rotatebox[origin=c]{90}{\textbf{Theory Solving}}}&
    \textbf{T-Backjump} & \( M~l^d~N~\parallel ~ F,~C  \) & \( \longrightarrow \) & \( M~ l' ~ \parallel~ F,~ C\) & 
    \textbf{if}\quad \(
    \begin{cases}
        Ml^d N \models \neg C, \text{ and } \exists ~C' \lor l'.\\
        (F,C \models_T C' \lor l') \land (M \models \neg C') \\
        l' \notin M\\
        l' \text{ or } \neg l' \text{ occurs in } F \text{ or in } Ml^dN\\
    \end{cases}
    \)\\\\
    &\textbf{T-Learn} & \( M \parallel~ F  \) & \( \longrightarrow \) & \( M~  \parallel~ F,~ C\) & 
    \textbf{if}\quad \(
    \begin{cases}
        \text{each atom of } C \text{ occurs in } F \text{ or } M\\
        F \models_T C
    \end{cases}
    \)\\\\    
    &\textbf{TheoryPropagate} & \( M \parallel~ F  \) & \( \longrightarrow \) & \( M~ l~ \parallel~ F \) & 
    \textbf{if}\quad \(
    \begin{cases}        
        l \notin M\\
        l \text{ or } \neg l \text{ occurs in } F\\
        M \models_T l\\        
    \end{cases}
    \)\\\\  
    \bottomrule{}
\end{tabular}}
\end{table}

Tab.~\ref{tab:rules} gives the conditional transition rules for \tool{}. \emph{Decision} literals, written with suffix $l^d$, are non-deterministically decided (i.e., guessed), while  other literals are deduced deterministically through implication. Intuitively, mistakes can happen with decision literals and require backtracking. In contrast, rules that add non-decision literals help prune the search space.

The rules \emph{Decide, BCP, Fail} describe transitions that do not rely on theory solving. \emph{Decide} non-deterministically selects and adds an undefined literal $l$ to $M$ (i.e., $l$ is a decision literal and can be backtracked to when conflict occurs). 
 \emph{BCP} (or UnitPropagate) infers and adds the unit literal $l$ to $M$ to satisfy the clause $C \lor l$, where $M \models \neg C$.
 \emph{Fail} moves to a \texttt{Fail} state (i.e., $F$ is unsatisfiable) when a conflicting clause $C$ occurs and $M$ contains no decision literals to backtrack to.

The rules \emph{T-Learn, T-Forget, T-Backjump, TheoryPropagate} describe transitions that rely on theory solving, e.g., $\models_T$. \emph{T-Backjump} analyzes a conflicting clause $C$ to determine an "incorrect" decision literal $l^d$ and computes a  "backjump" clause $C' \lor l'$ (which will be used by \emph{T-learn} to ensure that the incorrect decision literal $l$ will not be added to $M$ in the future).  The rule also adds $l'$ to $M$ (since $M \models \neg C'$) and removes $l^d$ and the set $N$ of subsequent literals added to $M$ after $l^d$ (i.e., it backtracks and removes the "incorrect" decision $l^d$ and subsequent assignments).
\emph{T-Learn} strengthens $F$ with a clause C that is entailed by $F$ (i.e., learned clauses are \emph{lemmas} of $F$). 
As mentioned, clause $C$ is the "backjumping" clause $C' \lor l'$ in \emph{T-Backjump}.
Finally, \emph{TheoryPropagate} infers literals that are T-entailed by literals in $M$ (thus $l$ is a non-decision literal).


\textbf{\tool{} Algorithm.}
The Decide and BCP rules align to the Decide and BCP components of \tool{}, respectively. 
The other rules are also implemented in \tool{} through the interactions of Deduction, Analyze-Conflict, and Backtrack components. For example, the T-Backjump rule is implemented as part of Deduction and AnalyzeConflict. Also note that while implication graph is a common way to detect conflicts and derive backjumping clause, it is still an implementation detail and therefore not mentioned in T-Backjump (which states there exists a way to obtain a backjumping clause).
T-Learn, which adds lemmas to existing clauses, is achieved in the main loop of the \tool{} algorithm (Fig.~\ref{fig:alg}, line~\ref{line:learn}). TheoryPropagate is implemented as part of \emph{Deduction} (Fig.~\ref{alg:deduction}, lines~\ref{line:infera}--\ref{line:inferb}). Finally, theory solving , i.e., $\models_T$, is implemented in Deduction by using LP solving and abstraction to check satisfiability of linear constraints.


 \subsection{Termination and Correctness of \tool{} DPLL(T)}

By describing \tool{} DPLL(T) using transition rules, we can now establish the the formal properties \tool{} DPLL(T), which are similar to those of standard DPLL(T).
Below we summarize the main results and refer the readers to~\cite{nieuwenhuis2006solving} for complete proofs.

Note that the work in~\cite{nieuwenhuis2006solving} covers multiple variants of DPLL with various rule configurations. Here we focus on just the base DPLL(T) algorithm of \tool{}. 
This significantly simplifies our presentation.



We first establish several invariants for the transition rules of \tool{} DPLL(T).

\begin{lemma}\label{lemma:a}
If $\emptyset \parallel F \Longrightarrow^* M \parallel G$, then the following hold: 

\begin{enumerate}
  \item All atoms in $M$ and all atoms in $G$ are atoms of $F$.
  \item $M$ is indeed an assignment, i.e., it contains no pair of literals $p$ and $\neg p$.
  \item $G$ is equivalent to $F$ in the theory T.
\end{enumerate}
\end{lemma}

All properties hold trivially in the initial state $\emptyset \parallel F$, so we will use induction to show the transition rules preserve them. Consider a transition $M'\parallel F' \Longrightarrow M'' \parallel F''$. Assume the properties hold for $M\parallel F$.
Property 1 holds because the only atoms can be added to $M''$ and $F''$ are from $M'$ and $F'$, all of which belong to $F$.
Property 2 preserves the requirement that $M$ never shares both negative and positive literals of an atom (the condition of each rule adding a new literal ensures this). 
Property 3 holds because only T-Learn rule can modify $F'$, but learning a clause $C$ that is a logical consequence  of $F'$ (i.e., $F' \models_T C$) will preserve the equivalence between $F'$ and $F''$.



\begin{lemma}\label{lemma:c}
If $\emptyset \parallel F \Longrightarrow^* S$, and $S$ is final 
 state, then $S$ is either Fail, or of the form $M \parallel F'$, where $M$ is a T-model of $F$.
\end{lemma}

This states that if $M\models F'$ then $M \models F$. This is true because $F$ and $F'$ are logical equivalence by Lemma~\ref{lemma:a}(3).

Now we prove that \tool{} DPLL(T) \textbf{terminates}.

\begin{theorem}[Termination]
Every derivation $\emptyset \parallel F \Longrightarrow S_1 \Longrightarrow \dots$ is finite.
\end{theorem}

This proof uses a well-founded strict partial ordering on \emph{states} $M \parallel F$. First, consider the case without T-Learn, in which only the assignment M is modified and the formula F remains constant. Then we can show no infinite derivation by (i) using Lemma~\ref{lemma:a}(1,2) that the number of literals in M and M' are always less than or equal to the number of atoms in F and (ii) show that the number of "missing" literals of M is strictly greater than those of M'. 
Now, consider the case with T-learn.  While F' can now be modified, i.e., learning new clauses, the number of possible clauses can be added to $F'$ is finite as clauses are formed from a finite set of atoms and the conditions of T-learn disallow clause duplication. 

Note that if \tool{} involved the Restart and Forget rules, which periodically remove learned clauses, then its termination argument becomes more complicated (but still holds) as shown in the work~\cite{nieuwenhuis2006solving}.

Now we prove that \tool{} DPLL(T) is \textbf{sound} and \textbf{complete}.

\begin{theorem}\label{theorem:soundness}
If $\emptyset \parallel F \Longrightarrow^* S$ where the state S is final, then 
\begin{enumerate}
\item \textbf{Sound}: S is Fail if, and only if, F is T-unsatisfiable
\item \textbf{Complete}: If $S$ is of the form $M \parallel F'$, then $M$ is a T-model of $F$.
\end{enumerate}
\end{theorem}

Property 1 states that \tool{} DPLL(T) ends at Fail state iff the problem F is unsatisfiable. 
Property 2 asserts that if \tool{} DPLL(T) ends with an assignment $M$, then $M$ is the model of $F$, i.e, $F$ is satisfiable. This property requires showing that if $M \models_T F'$, then $M \models_T F$, which is established in Lemma~\ref{lemma:c}.


Together, these properties of soundness, completeness, and termination make \tool{} DPLL(T) a decision procedure. Note that the presented results are independent from the theory under consideration. The main requirement of T-solver is its decidability  for T-satisfiability or T-consistency checking.
\tool{} uses LRA, a theory of real numbers with linear constraints, including linear equalities and inequalities, which is decidable~\cite{kroening2016decision}.

\end{document}